\documentclass[journal]{IEEEtran}
\usepackage{cite}
\usepackage{amsmath,amssymb,amsfonts}
\usepackage{amsthm}
\usepackage{mathrsfs}
\usepackage{algorithmic}
\usepackage{float}
\usepackage{graphicx}
\usepackage{subfigure}
\usepackage[justification=centering]{caption}
\usepackage{textcomp}
\usepackage{xcolor}
\usepackage{colortbl,booktabs}
\usepackage[ruled,vlined]{algorithm2e} 
\usepackage{times}  
\usepackage{helvet}  
\usepackage{courier}  
\usepackage{soul}
\usepackage{bm}
\usepackage[switch]{lineno}
\usepackage{pdfpages}
\newtheorem{theorem}{\textbf{Theorem}}
\newtheorem{theorem*}{Theorem}
\newtheorem{corollary}{Corollary}[theorem]

\newtheorem{Definition}{Definition}

\newtheorem{definition*}{Problem}

\newtheorem{Remark}{Remark}
\DeclareMathOperator*{\argmin}{arg\,min}

\usepackage{ifpdf}

\ifCLASSINFOpdf
\else
\fi
\usepackage{array}
\usepackage{fixltx2e}
\begin{document}
%
\title{Open Set Domain Adaptation: \\Theoretical Bound and Algorithm}
%
%
%
\author{Zhen Fang,~Jie~Lu,~\IEEEmembership{Fellow,~IEEE}, Feng Liu,~\IEEEmembership{Student Member,~IEEE},~Junyu Xuan,~and~Guangquan~Zhang
\thanks{Zhen Fang, Jie Lu, Feng Liu, Junyu Xuan and Guangquan Zhang are with the Centre for Artificial Intelligence, Faulty of Engineering and Information Technology, University of Technology Sydney,
NSW, 2007, Australia, e-mail: Zhen.Fang@student.uts.edu.au, \{ Jie.Lu; Feng.Liu; Junyu.Xuan; Guangquan.Zhang\}@uts.edu.au. }
}
\maketitle

\begin{abstract}
 The aim of unsupervised domain adaptation is to leverage the knowledge in a labeled (source) domain to improve  a model's learning performance with an unlabeled (target) domain -- the basic strategy being to mitigate the effects of discrepancies between the two distributions. Most existing algorithms can only handle unsupervised closed set domain adaptation (UCSDA), i.e., where the source and target domains are assumed to share the same label set. In this paper, we target a more challenging but realistic setting: unsupervised open set domain adaptation (UOSDA), where the target domain has unknown classes that are not found in the source domain. This is the first study to provide a learning bound for open set domain adaptation, which we do by theoretically investigating the risk of the target classifier on unknown classes. The proposed
learning bound has a
special term, namely open set difference, which reflects the risk
of the target classifier on unknown classes. Further, we present a novel and theoretically guided unsupervised algorithm for open set domain adaptation, called \textit{Distribution Alignment with Open Difference} (DAOD), which is based on regularizing this open set difference bound. The experiments on several benchmark datasets show the superior performance of the proposed UOSDA method compared with the state-of-the-art methods in the literature.
\end{abstract}


%
\IEEEpeerreviewmaketitle

\section{Introduction}
%
%
%
%

\IEEEPARstart{S}{tandard} supervised learning relies on the assumption that both the training and the testing samples are drawn from the same distribution. Unfortunately, this assumption does not hold in many applications since the process of collecting samples is prone to dataset bias \cite{candela2009dataset,DBLP:journals/kbs/LuBHZXZ15}. In object recognition, for example, there can be a discrepancy in the distributions between training and testing as a result of the given conditions, the device type, the position, orientation, and so on. To address this problem, \emph{unsupervised domain adaptation} (UDA) \cite{pan2008transfer,pan2010survey} has been proposed as a way of transferring relevant knowledge from a source domain that has an abundance of labeled samples to an unlabeled domain (the target domain). 

The aim of UDA is to minimize the discrepancy between the distributions of two domains. Existing work on UDA falls into two main categories: (1) feature matching, which seeks a new feature space where the marginal distributions or conditional distributions from the two domains are similar \cite{pan2011domain,long2013transfer,long2014transfer}, and (2) instance reweighting, which estimates the weights of the source domain so that the distributional discrepancy is minimized \cite{huang2007correcting,yu2012analysis}. There is an implicit \textit{assumption} in most existing UDA algorithms \cite{DBLP:journals/tnn/WeiKG19,DBLP:journals/tnn/LiSH17,DBLP:journals/tnn/PassalisT19,DBLP:conf/ijcnn/Fang00019,DBLP:journals/tfs/LiuLZ18,DBLP:journals/tnnls/LiuLZ20,DBLP:journals/tfs/ZuoZPBL17}  that {the source and target domains share the same label set}. Under this assumption, UDA is also regarded as \textit{unsupervised closed set domain adaptation} (UCSDA) \cite{DBLP:conf/iccv/BustoG17}. 

\begin{figure}[!]
\centering
\includegraphics[scale=0.34,trim=73 40 0 20, clip]{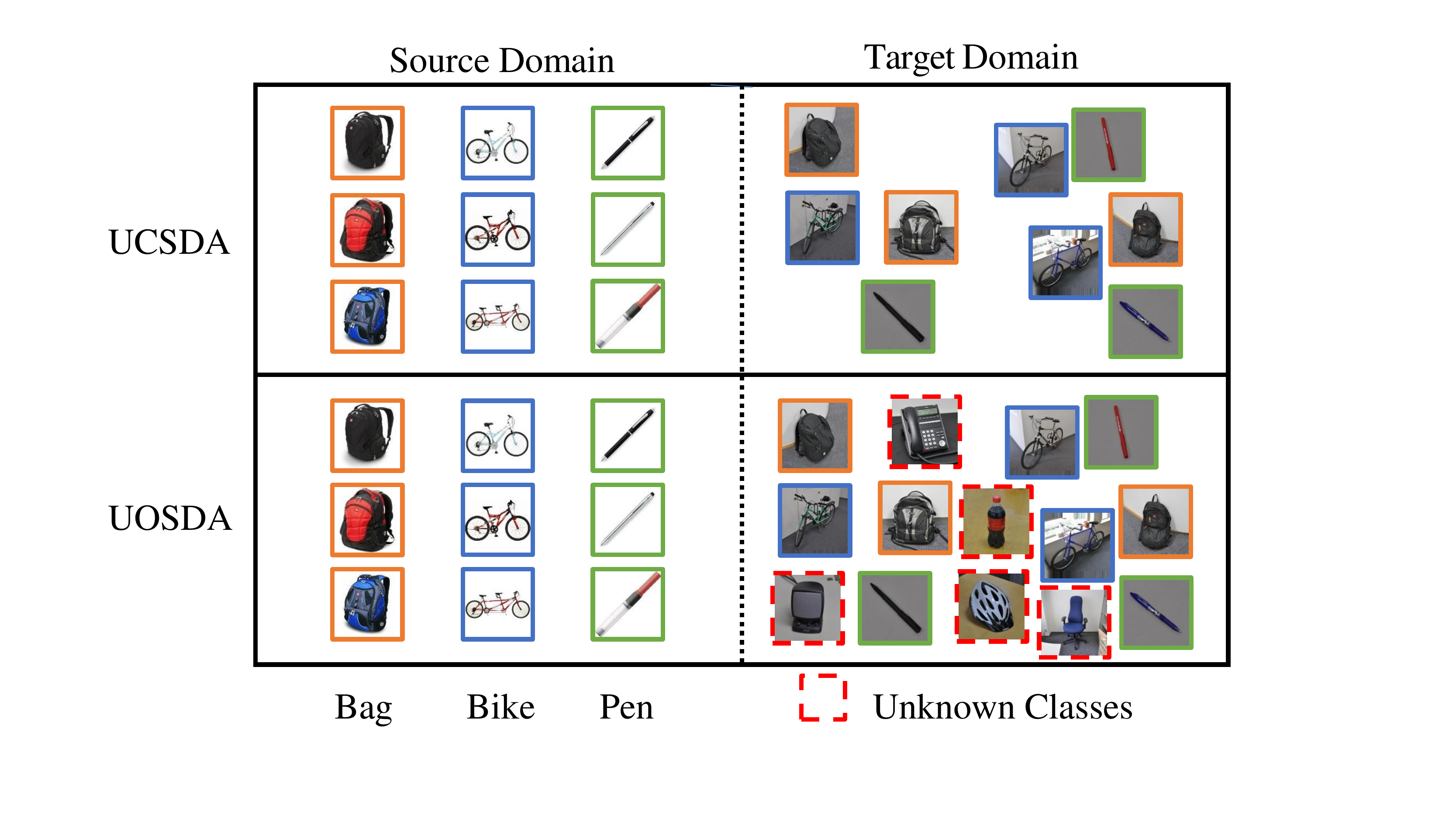}
\caption{Unsupervised open set domain adaptation problem (UOSDA), where the target domain contains ``unknown'' classes that are not contained in the label set of the source domain.
}
\end{figure}
\begin{figure}[t]
\includegraphics[scale=0.5,trim=250 130 0 70, clip]{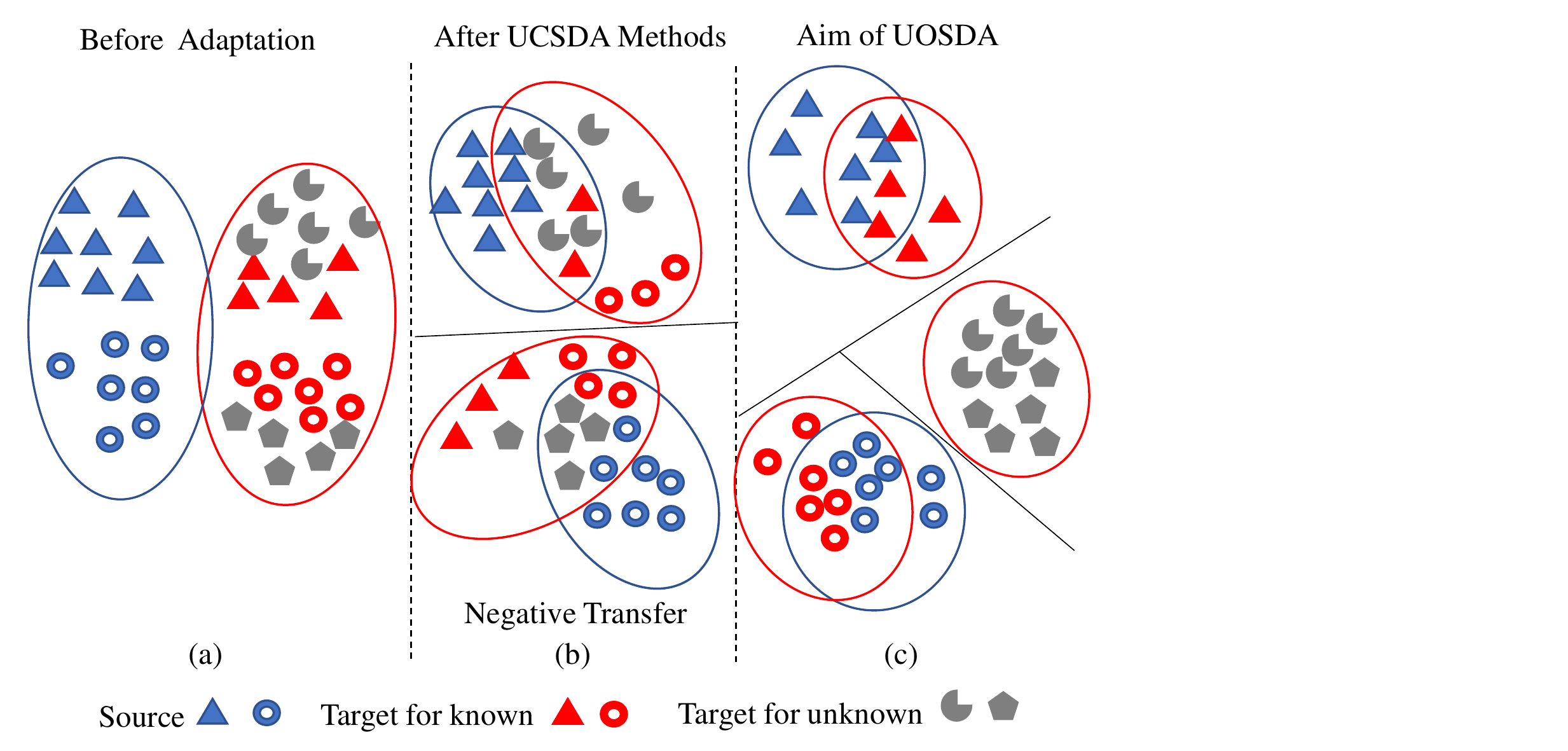}
\caption{Aim of UOSDA. (a) The original source and target samples are given. (b) UCSDA algorithm matches the source and target samples, leading to negative transfer. Because the unknown target samples interfere with distribution matching. (c) UOSDA algorithm classifies  known  target  samples  into  the correct  known classes and recognizes the unknown target samples as unknown.
}\label{fig:UOSDA}
\end{figure}

However, this assumption in UCSDA algorithms is not realistic in an unsupervised setting (i.e., there are no labels in the target domain) since it is not known whether the classes
of target samples are from the label set of the source domain. It may be that the target domain contains additional classes (\textit{unknown classes}) that do not exist in the label set of the source domain \cite{DBLP:conf/eccv/SaitoYUH18}. For example, in the Syn2Real task \cite{DBLP:journals/corr/abs-1806-09755}, there may be more classes for the real-world objects in the target domain than the synthetic objects contained in the source domain. Therefore, if existing UCSDA algorithms were to be used to solve the UDA problem without the assumption in UCSDA, the potential mismatches between unknown and known classes would likely result in negative transfer \cite{rosenstein2005transfer} (see Fig. \ref{fig:UOSDA}(b)).

To address UDA problem \textit{without} the assumption, Busto et al. \cite{DBLP:conf/iccv/BustoG17} and Saito et al. \cite{DBLP:conf/eccv/SaitoYUH18}
recently proposed a new problem setting, \textit{unsupervised open set domain adaptation} (UOSDA), in which the unlabeled target domain contains unknown classes that do not belong to the label set of the source domain (see Fig. 1). 
There are two key challenges \cite{DBLP:conf/eccv/SaitoYUH18} in addressing the UOSDA problem. 
The first challenge is that there is not enough knowledge in the target domain to classify the unknown samples. So how should these samples be labeled? The solution is to mine deeper information in the target domain to delineate a boundary
between the known and unknown classes. The second challenge in UOSDA is the difference in distributions. The unknown target samples should not be matched when the overall distribution is matched, otherwise negative transfer may occur. 

Only a small number of algorithms have been proposed to solve the UOSDA problem \cite{DBLP:conf/iccv/BustoG17,DBLP:conf/eccv/SaitoYUH18,baktashmotlagh2018learning,longmingsheng,DBLP:conf/icmcs/ZhangLHCZG19}. The first proposed UOSDA algorithm is \textit{Assign-and-Transform-Iteratively} (ATI) \cite{DBLP:conf/iccv/BustoG17}, which recognizes unknown target samples using constraint integer programming. It then learns a linear map to match the source domain with the target domain by excluding the predicted unknown target samples. However, ATI carries the assumption that the source domain contains unknown classes that are not in the target domain. Hence, the first proposed deep UOSDA algorithm,  \textit{Open  Set  Back Propagation}  (OSBP) \cite{DBLP:conf/eccv/SaitoYUH18}  was developed to address the UOSDA problem without this assumption. It rejects unknown target samples by training a binary cross entropy loss.  

Although ATI and OSBP are designed to solve the UOSDA problem, neither is based on a theoretical analysis of UOSDA. Moreover, no work has yet given a learning bound for open set domain adaptation problems. To fill this gap, this paper presents a theoretical exploration of UOSDA. In studying the risk of the target classifier on unknown classes, we discovered the risk is closely related to a special term called \textit{open set difference} which can be estimated from the unlabeled samples. Minimizing the open set difference helps us to classify unknown target samples, addressing the first challenge.

Following our theory, we design a principle-guided UOSDA algorithm referred to as \textit{Distribution Alignment with Open Difference} (DAOD). This algorithm can accurately classify unknown target samples while minimizing the discrepancy between the two domains for known classes. DAOD learns the target classifier by simultaneously optimizing the structural risk function \cite{DBLP:books/daglib/0097035}, the joint distribution alignment, the manifold regularization \cite{DBLP:journals/jmlr/BelkinNS06}, and open set difference.  The reason DAOD is able to avoid negative transfer lies in its ability to minimize the open set difference, which enables the unknown target samples to be classified accurately as unknown. By excluding these recognized unknown target samples, the source and target domains can be precisely aligned, addressing the second challenge.





As mentioned, there is no theoretical work in the literature for open set domain adaptation. The closest theoretical work is by Ben-David et al. \cite{DBLP:conf/nips/Ben-DavidBCP06}, who gives VC-dimension-based generalization bounds. Unfortunately, this work has several restrictions: 1) the theoretical analysis only covers closed settings; and 2) the work only solves binary classification tasks, rather than the multi-class problems common to open settings. A significant contribution of this paper is that the theoretical work gives a learning bound for open set domain adaptation. 


The contributions of this paper are summarized as
follows.

$\bullet$  We provide the theoretical bound  for open set domain adaptation. The closed set domain adaptation theory \cite{DBLP:conf/nips/Ben-DavidBCP06} is a special case of our theoretical results. To the best of our
knowledge, this is the first work on open set domain adaptation theory. 


$\bullet$ We develop an unsupervised novel open set domain adaptation algorithm, Distribution Alignment with Open Difference (DAOD), which is based on the open set learning bound proposed.  The algorithm enables  the unknown target samples to be separated from samples using open set difference. 

$\bullet$  We conduct $38$ real-world UOSDA tasks (including $20$ face recognition tasks and $18$ object recognition tasks) for evaluating DAOD and existing UOSDA algorithms. Extensive experiments demonstrate that DAOD outperforms the state-of-the-art UOSDA algorithms ATI and OSBP.  

This paper is organized as follows. Section II reviews existing work on unsupervised closed set domain adaptation, open set recognition and unsupervised open set domain adaptation. Section III presents the definitions, important notations and our problem. Section IV provides the main theoretical result and our proposed algorithm. Comprehensive evaluation results and analyses are provided in Section V. Lastly, Section VI concludes the paper. 


\section{Related Work}
In this section, we present relevant work related to unsupervised closed set domain adaptation algorithms, open set recognition, and unsupervised open set domain adaptation.

\textbf{Closed Set Domain Adaptation.} Ben-David et al.\cite{DBLP:conf/nips/Ben-DavidBCP06} proposed learning bounds for closed set domain adaptation, where the bounds show that the performance of the target classifier depends on the performance of the source classifier and the discrepancy between the source and target domains. Many UCSDA algorithms \cite{long2014transfer,DBLP:journals/tnn/LiSH17,DBLP:journals/tnn/DengLOTCZ19,liu2019butterfly} have been proposed based on theoretical bounds with the objective of minimizing the discrepancy between domains. These algorithms can be roughly divided into two categories: feature matching and instance reweighting.

Feature matching aims to reduce the distribution discrepancy by learning a new feature representation. \textit{Transfer Component Analysis} (TCA) \cite{pan2011domain} learns a new feature space to match distributions by employing the \textit{Maximum Mean Discrepancy} (MMD) \cite{gretton2012kernel}. \textit{Joint Distribution Adaptation} (JDA) \cite{long2013transfer} improves TCA by jointly matching marginal distributions and conditional distributions. \textit{Adaptation Regularization
Transfer Learning} (ARTL) \cite{DBLP:journals/tkde/LongWDPY14} considers a manifold regularization term \cite{DBLP:journals/jmlr/BelkinNS06} to learn the geometric relations between domains, while matching distributions. \textit{Joint Geometrical and Statistical Alignment} (JGSA) \cite{DBLP:conf/cvpr/ZhangLO17} not only considers the distribution discrepancy but also matches the geometric shift. Recent advances show that deep networks can be successfully applied to closed set domain adaptation tasks.  \textit{Deep Adaptation Networks} (DAN) \cite{long2015learning} considers three adaptation layers for matching distributions and applies Multiple Kernels MMD \cite{gretton2012optimal} for adapting deep representations. \textit{Wasserstein Distance Guided Representation Learning} \cite{shen2018wasserstein} minimizes the distribution discrepancy by employing \textit{Wasserstein Distance} in neural networks. 

In the other category, instance reweighting algorithms reduce the distribution discrepancy by weighting samples in the source domain. \textit{Kernel Mean Matching} \cite{huang2007correcting} defines the weights as the density ratio between the source domain and the target domain. Yu et al. \cite{yu2012analysis} has provided a theoretical analysis for important instance reweighting algorithms. However, with a very great domain discrepancy, a large number of effective source samples are down-weighted and useful information is lost.

Unfortunately, the algorithms mentioned above cannot be applied to open set domain adaptation because unknown target samples would be included in the distribution matching process, leading to negative transfer.

\textbf{Open Set Recognition.} When the source domain and target domain for known classes share the same distribution, open set domain adaptation becomes \textit{Open Set Recognition}. A common method for handling open set recognition relies on the use of threshold-based classification strategies \cite{DBLP:books/daglib/p/PhillipsGM11}. Establishing a threshold for the similarity score means distant samples are removed from the training samples.  \textit{Open Set Nearest Neighbor} (OSNN) \cite{DBLP:journals/ml/Mendes-JuniorSW17} recognizes whether a sample is from an unknown class by comparing the threshold with a ratio: the similarity score of the sample to the two classes most similar to that sample. Another research stream relies on modifying \textit{Support Vector Machines}  \cite{DBLP:journals/pami/ScheirerJB14,DBLP:conf/icb/VaretoSCS17,DBLP:conf/eccv/JainSB14}. \textit{Multi-class open set SVM} \cite{DBLP:conf/eccv/JainSB14} uses a multi-class SVM as a basis to learn the unnormalized posterior probability which is used to reject unknown samples.

\textbf{Open Set Domain Adaptation.} The open set domain adaptation problem was proposed by Assign-and-Transform-Iteratively (ATI) \cite{DBLP:conf/iccv/BustoG17}. Using $\ell_2$ distance between each target sample and the center of each source class, ATI constructs a constraint integer programming to recognize unknown target samples $S_u$, then learns a linear transformation to match the source domain and  target domain excluding $S_u$. However, ATI requires the help of unknown source samples, which are unavailable in our setting. Recently, a deep learning algorithm, Open Set Back Propagation (OSBP) \cite{DBLP:conf/eccv/SaitoYUH18}, is a recent contribution to addressing UOSDA. OSBP relies on an adversarial neural network and a binary cross entropy loss to learn the probability of the target samples. It then uses the estimated probability to separate samples of known and unknown classes in the target. However, we have not found any paper that considers the learning bound for open set domain adaptation. In this paper, we aim to fill in the blanks of open set domain adaptation theory.
 \section{ {Preliminaries}}
In this section, we formally define the problem setting for this paper and introduce some fundamental concepts to domain adaptation and, therefore, this study. The notations used throughout this paper are summarized in Appendix A. Table I .

\subsection{Definitions and Problem Setting}

 Important definitions are presented as follows.
\begin{Definition}[{Domain}]\label{d2} Given a feature (input) space $\mathcal{X}\subset \mathbb{R}^d$ and a label (output) space $\mathcal{Y}$, a \textit{domain} is a joint distribution $P(X, Y)$, where random variables $X \in \mathcal{X}$, $Y \in \mathcal{Y}$.
\end{Definition}

To be exact, a random variable is a measurable map. In Definition 1, $X\in \mathcal{X}$ and  $Y \in \mathcal{Y}$ mean that the image sets of $X$ and $Y$ are contained in the spaces $\mathcal{X}$ and $\mathcal{Y}$ respectively.
We normally name the random variable $X$ from the feature space $\mathcal{X}$ as feature vector while the random variable $Y$ as label. The label $Y$ can either be continuous (in a regression task) or discrete (in a classification task). In this paper, we have fixed it as a discrete variable with a fixed number of items. Based on this definition, we have:
\begin{Definition}[{Domains for Open Set Domain Adaptation}]\label{d3}Given a feature space $\mathcal{X}\subset \mathbb{R}^d$ and the label spaces $\mathcal{Y}^s, \mathcal{Y}^t$, the {source} and {target domains} have different joint distributions $P(X^s, Y^s)$ and $P(X^t, Y^t)$, where the label space $\mathcal{Y}^s \subset \mathcal{Y}^t$, and  random variables $X^s, X^t \in \mathcal{X}$, $Y^s \in \mathcal{Y}^s$, $Y^t \in \mathcal{Y}^t$.
\end{Definition}

From Definitions 1 and 2, we can see that: 1) $X^s$ and $X^t$ are from the same space because our focus is on homogeneous situations; and 2) $\mathcal{Y}^s$ is a subset of $\mathcal{Y}^t$. The classes from $\mathcal{Y}^t \backslash \mathcal{Y}^s$ are the \textit{{unknown target classes}}.  The classes from $\mathcal{Y}^s$ are the \textit{{known classes}}. Thus, the problem to be solved is:

\begin{definition*}[{Unsupervised Open Set Domain Adaptation (UOSDA)}]
Given labeled samples $\mathcal{S}$ drawn from the source domain $P(X^s, Y^s)$ i.i.d. and unlabeled samples $\mathcal{T}_X$ drawn from the target marginal distribution $P(X^t)$ i.i.d., {the aim of unsupervised open set domain adaptation} is to find a target classifier  $f^t:\mathcal{X}\rightarrow \mathcal{Y}^{t}$ such that
\newline
\noindent\textit{1)} $f^t$ \textit{classifies known target samples into the correct known classes;}\newline
\noindent\textit{2)} $f^t$ \textit{classifies unknown target samples as unknown.
}
\end{definition*}

It is worth noting that, with UOSDA tasks, the algorithm only needs to classify unknown samples as unknown and classify known target samples into the correct known classes. Classifying unknown target samples into correct unknown classes is not necessary. Hence, we consider all unknown target samples are allocated to one big unknown class.  {Without loss of generality, we assume that
$\mathcal{Y}^s=\{{\mathbf{y}}_c\}_{c=1}^C,$ $\mathcal{Y}^t=\{{\mathbf{y}}_c\}_{c=1}^{C+1}$, where the label ${\mathbf{y}}_{C+1}$ represents the unknown target classes and the label ${\mathbf{y}}_c\in \mathbb{R}^{{(C+1)}\times 1}$ is a one-hot vector, whose $c$-th coordinate is $1$ and other coordinates are $0$. The label ${\mathbf{y}}_c$ represents the $c$-th class.}

\subsection{{Concepts and Notations}}
\vspace{-0.1cm}
Before introducing our main results, we need to introduce the following necessary concepts and notations in this field. 
Unless otherwise specified, all the following notations are used consistently throughout the paper without further explanations. More detail on these notations is provided in Appendix A.

\subsubsection{{Notations for distributions}} For the
sake of simplicity, we use the notations $P_{X^sY^s}$ and $P_{X^tY^t}$ to denote the joint distributions $P(X^s,Y^s)$ and $P(X^t,Y^t)$ respectively and also denote $P_{X^s}$ and $P_{X^t}$ as the marginal distributions $P(X^s)$ and $P(X^t)$, respectively. 

$P_{X^s|{\mathbf{y}}_c}$ and $P_{X^t|{\mathbf{y}}_c}$ represent the conditional distributions for the $c$-th class $P(X^s|Y^s={\mathbf{y}}_{c})$ and $P(X^t|Y^t={\mathbf{y}}_{c})$, while $\pi^t_c$ represents the {target class-prior probability} for $c$-th class $P(Y^t=\mathbf{y}_{c})$. Hence, $\pi^t_{C+1}=P(Y^t={\mathbf{y}}_{C+1})$ is the {class-prior probability for the unknown target classes}.

Lastly, $P_{X^t|\mathcal{Y}^s}$ represents the target conditional distribution for the known classes $P({X^t|Y^t\in \mathcal{Y}^s})$, which can be evaluated by
\begin{equation*}
\frac{ P(X^t,Y^t \in \mathcal{Y}_s)}{P(Y^t \in \mathcal{Y}_s)}=\frac{\sum_{c=1}^C P(X^t|Y^t = \mathbf{y}_c)\pi^t_{c}}{1-\pi^t_{C+1}}.
\end{equation*}
The notation $\widehat{P}$ denotes the corresponding empirical distribution to any distribution $P$. For example, $\widehat{{P}}_{X^sY^s}$ represents the empirical distribution corresponding to ${{P}}_{X^sY^s}$.

\subsubsection{{Risks and Partial Risks}} Risks and partial risks are two important concepts in learning theory, which are briefly explained in the following and later used in our theorems. 

Following the notations in \cite{DBLP:conf/icml/0002LLJ19}, we consider a
multi-class classification task with a {\textit{hypothesis space}} $\mathcal{H}$ of the {\textit{scoring functions}}
\begin{equation}\label{hyp}
\begin{split}
    ~~~~~~~{\bm {h}}:~\mathcal{X}&\rightarrow \mathbb{R}^{|\mathcal{Y}^t|}=\mathbb{R}^{C+1}\\
    {\mathbf{x}}&\rightarrow [h_1({\mathbf{x}}),...,h_{C+1}({\mathbf{x}})]^T,
    \end{split}
\end{equation}
where the output $h_c({\mathbf{x}})$ indicates the confidence in the prediction of the label $\mathbf{ y}_c$. Let $\ell: \mathbb{R}^{C+1} \times \mathbb{R}^{C+1} \rightarrow \mathbb{R}_{+}$ be a {\textit{symmetric loss function}}.
Then the {\textit{risks}} of ${\bm h}\in \mathcal{H}$ w.r.t. $\ell$ under ${P}_{X^sY^s}$ and $P_{X^tY^t}$  are given by 
\begin{equation}\label{kk}
   \begin{split}
      R^s({\bm h}):&=\underset{{({\mathbf{x}},{\mathbf{y}})\sim{P}_{X^sY^s}}}{\mathbb{E}}\ell({\bm h}({\mathbf{x}}),{\mathbf{y}})=\mathbb{E}~\ell({\bm h}(X^s),Y^s)),\\R^t({\bm h}):&=\underset{{({\mathbf{x}},{\mathbf{y}})\sim{P}_{X^tY^t}}}{\mathbb{E}}\ell({\bm h}({\mathbf{x}}),{\mathbf{y}})=\mathbb{E}~\ell({\bm h}(X^t),Y^t)).
  \end{split}
\end{equation}

The {\textit{partial risk}} of ${\bm h}\in \mathcal{H}$ for the known target classes is
\begin{equation}\label{k1}
\begin{split}
R_*^t({\bm h}):&=\frac{1}{1-\pi^t_{C+1}}\int_{\mathcal{X}\times \mathcal{Y}^s}\ell({\bm h}({\mathbf{x}}),{\mathbf{y}}){\rm d}P_{X^tY^t}({\mathbf{x}},{\mathbf{y}})
\end{split}
\end{equation}
and the {\textit{partial risk}} of  ${\bm h}\in \mathcal{H}$ for the unknown target classes is
\begin{equation}\label{-k}
\begin{split}
   {R}_{C+1}^t({\bm h}):&=\underset{{{\mathbf{x}}\sim P_{X^t|{\mathbf{y}}_{C+1}}}}{\mathbb{E}} \ell({\bm h}({\mathbf{x}}),{\mathbf{y}}_{C+1})\\&=\int_{\mathcal{X}}\ell({\bm h}({\mathbf{x}}),{\mathbf{y}}_{C+1}){\rm d}P_{X^t|{\mathbf{y}}_{C+1}}(\mathbf{x}).
    \end{split}
\end{equation}

According to (\ref{kk}), (\ref{k1}) and (\ref{-k}), we have
\begin{equation}\label{another}
   \begin{split}
     R^t({\bm h})=\pi^t_{C+1}R^t_{C+1}({\bm h})+(1-\pi_{C+1}^t)R_*^t({\bm h}).
     \end{split}
\end{equation}
 The proof can be found in Appendix A.

Lastly, we denote  
\begin{equation}\label{10000}
\begin{split}
&R^s_{u,C+1}({\bm h}):=\underset{{\mathbf{x}}\sim P_{X^s}}{\mathbb{E}}\ell({\bm h}({\mathbf{x}}),{\mathbf{y}}_{C+1})=\mathbb{E}~\ell({\bm h}(X^s),{\mathbf{y}}_{C+1}),\\&
R^t_{u,C+1}({\bm h}):=\underset{{{\mathbf{x}}\sim P_{X^t}}}{\mathbb{E}}\ell({\bm h}({\mathbf{x}}),{\mathbf{y}}_{C+1})=\mathbb{E}~\ell({\bm h}(X^t),{\mathbf{y}}_{C+1})
\end{split}
\end{equation}
as the {\textit{risks}} that the samples are regarded as the unknown classes. 

Given a risk $R({\bm h})$, it is convenient to use notation $\widehat{R}({\bm h})$ as the empirical risk that corresponds to $R({\bm h})$. Hence, notations $\widehat{{R}}^s({\bm h})$, $\widehat{{R}}^s_{u,C+1}({\bm h})$ and $\widehat{R}^t_{u,C+1}$ represent the empirical risks corresponding to the risks ${{R}}^s({\bm h})$, ${{R}}^s_{u,C+1}({\bm h})$ and ${R}^t_{u,C+1}({\bm h})$ respectively.
\subsubsection{{Discrepancy Distance and Maximum Mean Discrepancy}}
One challenge of domain adaptation is the mismatch between the distributions of the source and target domains. To mitigate this effect, two famous distribution distances have been proposed as the measures of the distribution difference. 

The first one is discrepancy distance presented as follows.
\begin{Definition}[{Discrepancy Distance}\cite{DBLP:conf/colt/MansourMR09}]\label{d0}Let the hypothesis space $\mathcal{H}$ be a set of functions defined in a feature space $\mathcal{X}$,  $\ell$ be a  loss function and $P_1,P_2$ be distributions on space $\mathcal{X}$. The {discrepancy~distance} $d_{\mathcal{H}}^{\ell}(P_1,P_2)$ between the distributions $P_1$ and $P_2$ over $\mathcal{X}$ is
\begin{equation*}
    \underset{{\bm h},{\bm h}^*\in \mathcal{H}}{\sup}|\underset{{{\mathbf{x}}\sim P_1}}{\mathbb{E}}\ell({\bm h}({\mathbf{x}}),{\bm h}^*({\mathbf{x}}))-\underset{{\mathbf{x}}\sim P_2}{\mathbb{E}}\ell({\bm h}({\mathbf{x}}),{\bm h}^*({\mathbf{x}}))|.
\end{equation*}
\end{Definition}

If $\ell$ in the definition is the zero-one loss, the discrepancy  distance is known as the $\mathcal{H}\Delta \mathcal{H}$ distance \cite{DBLP:conf/nips/Ben-DavidBCP06}. The discrepancy distance is symmetric and satisfies the triangle inequality,
but it does not define a distance in general: $d_{\mathcal{H}}^{\ell}(P_1,P_2)=0$ does not mean $P_1=P_2$.

The second distance is Maximum Mean Discrepancy: 
\begin{Definition}[{Maximum Mean Discrepancy} \cite{gretton2012kernel}]\label{MMD}
Given a feature space $\mathcal{X}$ and a class of function $\mathcal{F}$ $(f:\mathcal{X}\rightarrow \mathbb{R})$. The {maximum
mean discrepancy} between the distributions $P_1$ and $P_2$ is
\begin{equation*}
    \textnormal{MMD}[\mathcal{F},P_1,P_2]:=\sup_{f\in \mathcal{F}}|\underset{{{\mathbf{x}}\sim P_1}}{\mathbb{E}}f({\mathbf{x}})-\underset{{\mathbf{x}}\sim P_2}{\mathbb{E}}f({\mathbf{x}})|.
\end{equation*}
\end{Definition}

To ensure that $\textnormal{MMD}$ is a metric, one must identify a function class ${\mathcal{F}}$ that is rich enough to uniquely identify whether $P_1=P_2$. Gretton et al. \cite{gretton2012kernel}, therefore, propose as $\textnormal{MMD}$ function class ${\mathcal{F}}$ the unit ball in a \textit{reproducing kernel Hilbert space} (RKHS) $\mathcal{H}_{{k}}$ \cite{ghifary2017scatter}  (the subscript $k$ represents the reproducing
kernel and is used to distinguish the hypothesis set $\mathcal{H}$ from the RKHS $\mathcal{H}_{{k}}$).

For convenience, we have used the notation $\textnormal{MMD}_{\mathcal{H}_k}(\cdot,\cdot)$ to represent $\textnormal{MMD}[\mathcal{F},\cdot,\cdot]$, when $\mathcal{F}=\{f\in \mathcal{H}_k:~\|f\|_{\mathcal{H}_k}\leq 1\}$ \cite{gretton2012kernel}.
Note that $\textnormal{MMD}_{\mathcal{H}_k}$  is symmetric and satisfies the triangle inequality. When the kernel $k$ is a \textit{universal kernel}, $\textnormal{MMD}_{\mathcal{H}_k}(P_1,P_2)=0$ if and only if $P_1=P_2$, which implies that $\textnormal{MMD}_{\mathcal{H}_k}$ is a metric.

Though the MMD distance is powerful, it is not convenient to be optimized as a regularization term in shallow domain adaptation algorithms. The \textit{projected MMD} \cite{pan2011domain,ghifary2017scatter,DBLP:conf/cikm/QuanzH09}  has been proposed to transform the MMD distance into a proper regularization term. Given a scoring function ${\bm h}=[h_1,...,h_{C+1}]^T$, where $h_c\in \mathcal{H}_k, c=1,...,C+1$, the projected MMD is defined as follows: 
\begin{equation*}
\begin{split}
    D_{{\bm h},k}(P_1,P_2)=\left\|\int_{\mathcal{X}}{\bm h}(\mathbf{x}){\rm d}P_1(\mathbf{x})-\int_{\mathcal{X}}{\bm h}(\mathbf{x}){\rm d}P_2(\mathbf{x})\right\|_2,
    \end{split}
\end{equation*}
where $\|\cdot\|_2$ is the $\ell_2$ norm.
\subsubsection{Manifold Regularization}
{The idea of manifold regularization has a rich machine learning history going back to transductive learning and truly semi-supervised learning} \cite{DBLP:journals/jmlr/BelkinNS06}. {Manifold regularization is specifically designed to control the
complexity as measured by the geometry of the distribution. Given samples $\{\mathbf{x}_1,...,\mathbf{x}_n\}$, the manifold regularization is}
\begin{equation*}
\begin{split}
    &\sum^{n}_{i,j=1}\|{\bm h}(\mathbf{x}_i)-{\bm h}(\mathbf{x}_{j})\|_2^2\mathbf{W}_{ij},
    \end{split}
\end{equation*}
{where $[\mathbf{W}_{ij}]$ is the pair-wise affinity matrix and $\mathbf{W}_{ij}$ estimates the similarity of $\mathbf{x}_i,\mathbf{x}_j$.}

{By  the  manifold  assumption} \cite{DBLP:journals/jmlr/BelkinNS06}, {if two samples from the support set of the distributions $P_{X^s},P_{X^t}$ are  close, then  the  scores  of  the  two samples are similar. To extract geometric relationship between domains, the manifold regularization has been used by many closed set domain adaptation algorithms} \cite{ DBLP:conf/cikm/QuanzH09,DBLP:journals/tkde/LongWDPY14,DBLP:conf/mm/WangFCYHY18,DBLP:journals/tkde/Long0DS014,DBLP:journals/kbs/ZhangLO19,DBLP:journals/pami/LiZ19,DBLP:conf/ijcai/WangM11}.

\section{Proposed Algorithm}
\subsection{Main Theoretical Result and Open Set Difference}
Before introducing the main theorem, we firstly define open set difference, one of the main contributions of the paper. 
\begin{Definition}[{Open Set Difference}]\label{OSD} Given risks $R^s_{u,C+1}({\bm h})$ and $R^t_{u,C+1}({\bm h})$ defined in (\ref{10000}), the open set difference is
\begin{equation*}
    \Delta_o= \frac{R^t_{u,C+1}({\bm h})}{1-\pi_{C+1}^t}
 -R^s_{u,C+1}({\bm h}),
\end{equation*}
where $\pi_{C+1}^t$ is the  class-prior probability for the unknown target classes.
\end{Definition}
The following
theorem provides an open set domain adaptation bound according to discrepancy distance and open set difference.
\begin{theorem}\label{-1000}
Given a symmetric loss function $\ell$ satisfying triangle inequality and  a hypothesis $\mathcal{H}$ with a mild condition that the constant vector value function ${\bm g}:={\mathbf{y}}_{C+1}\in \mathcal{H}$, 
then for any ${\bm h}\in \mathcal{H}$,  we have 
\begin{equation*}
\begin{split}
 ~~\frac{R^t({\bm h})}{1-\pi_{C+1}^t}&\leq   \overbrace{R^s({\bm h})}^{\text{{Source Risk}}} +{2\overbrace{d_{\mathcal{H}}^{\ell}(P_{X^t|\mathcal{Y}^s},P_{X^s})}^{\text{{Distribution Discrepancy}}}}+\Lambda\\&+\underbrace{\frac{R^t_{u,C+1}({\bm h})}{1-\pi_{C+1}^t}
 -R^s_{u,C+1}({\bm h})}_{\text{{Open Set Difference}} \Delta_o},
 \end{split}
\end{equation*}
 where $R^s({\bm h})$ and $R^t({\bm h})$ are the risks defined in (\ref{kk}), $R^s_{u,C+1}({\bm h})$ and $R^t_{u,C+1}({\bm h})$ are the risks defined in (\ref{10000}), $R^t_{*}({\bm h})$ is the partial risk defined in (\ref{k1}), and $\Lambda=\underset{{\bm h}\in \mathcal{H}}{\min}  ~R^s({\bm h})+R^t_{*}({\bm h})$.
\end{theorem}
\begin{proof}
{Here we provide a proof sketch. Detailed proof is given in Appendix A.} {According to} (\ref{another}), {we have}
\begin{equation}\label{Th1}
\begin{split}
    &\frac{R^t({\bm h})}{1-\pi_{C+1}^t}-R^s({\bm h})\\=&R^t_*({\bm h})-R^s({\bm h})+\frac{\pi_{C+1}^t}{1-\pi_{C+1}^t}R^t_{C+1}({\bm h}).
\end{split}
\end{equation}
{Then we can check that }
\begin{equation}\label{Th2}
   R^t_{*}({\bm h})-R^s({\bm h})\leq\Lambda +d_{\mathcal{H}}^{\ell}(P_{X^t|\mathcal{Y}^s},P_{X^s}), 
\end{equation}
 \begin{equation}\label{Th3}
 \begin{split}
     \frac{\pi_{C+1}^t}{1-\pi_{C+1}^t}R^t_{C+1}({\bm h})\leq  {d_{\mathcal{H}}^{\ell}(P_{X^t|\mathcal{Y}^s},P_{X^s})}+\Delta_o.
     \end{split}
 \end{equation}

{Combining}  (\ref{Th2}),  (\ref{Th3}) {with} (\ref{Th1}), {we have}
 \begin{equation*}
    \frac{R^t({\bm h})}{1-\pi_{C+1}^t}\leq   {R^s({\bm h})} +{2{d_{\mathcal{H}}^{\ell}(P_{X^t|\mathcal{Y}^s},P_{X^s})}}+\Lambda+\Delta_o.
\end{equation*}
\end{proof}
{{\Remark{{The condition ${\mathbf{y}}_{C+1}\in \mathcal{H}$ can be replaced by a weaker condition that there exists a sequence  $\{{\bm h}_i\}_{i=1}^{+\infty}$  such that ${\bm h}_i$ converges uniformly to $\mathbf{y}_{C+1}$. Note that the hypothesis space $\mathcal{H}$ used in our algorithm satisfies the weaker condition automatically, thus, the condition ${\mathbf{y}}_{C+1}\in \mathcal{H}$ can be removed when we use the $\mathcal{H}$ 
applied in our algorithm. }}}}

  The open set difference $\Delta_o$ is the crucial term to bound the risk of ${\bm h}$ on unknown target classes, since
\begin{equation}\label{EQ}
    R^t_{C+1}({\bm h})\leq \frac{1-\pi_{C+1}^t}{\pi_{C+1}^t}\left(\Delta_o+d_{\mathcal{H}}^{\ell}(P_{X^t|\mathcal{Y}^s},P_{X^s})\right).
\end{equation}
The risk of ${\bm h}$ on unknown target classes is intimately linked to the open set difference $\Delta_o$:
\begin{equation*}
    \left|\pi_{C+1}^tR^t_{C+1}({\bm h})- ({1-\pi_{C+1}^t})\Delta_o
    \right| \leq d_{\mathcal{H}}^{\ell}(P_{X^t|\mathcal{Y}^s},P_{X^s}).
\end{equation*}
When $\pi_{C+1}^t=0$, {Theorem \ref{-1000}} degenerates into the  closed set scenario with this theoretical bound \cite{DBLP:conf/nips/Ben-DavidBCP06}
\begin{equation*}
R^t({\bm h})\leq   {R^s({\bm h})} +{3{d_{\mathcal{H}}^{\ell}(P_{X^t}},P_{X^s})+\Lambda}.
\end{equation*}
This is because when $\pi_{C+1}^t=0$, the open set difference is
\begin{equation*}
\Delta_o\leq d_{\mathcal{H}}^{\ell}({P}_{X^t|\mathcal{Y}^s},P_{X^s})=d_{\mathcal{H}}^{\ell}(P_{X^t},P_{X^s}).
\end{equation*} 
The significance of {Theorem \ref{-1000}} is twofold. First, it highlights that the open set difference  $\Delta_o$ is the main term for controlling performance in open set domain adaptation. Second, the bound shows a direct connection with closed set domain adaptation theory. 

In addition, the open set difference $\Delta_o$ consists of two parts: a positive term $R^t_{u,C+1}({\bm h})$ and a negative term $R^s_{u,C+1}({\bm h})$. A larger positive term implies more target samples are classified as unknown samples. The negative term is used to prevent the source samples from being classified as unknown. According to (\ref{EQ}),
  the negative term and the distance discrepancy jointly prevent all target samples from being recognized as unknown classes.  In addition, {Corollary \ref{0.1}} also tells us that the positive term and the negative term can be estimated from unlabeled samples. Using \textit{Natarajan Dimension Theory} \cite{DBLP:books/mk/Natarajan91} to bound the source risk $R^s({\bm h})$, risks $R^t_{u,C+1}({\bm h})$ and $R^s_{u,C+1}({\bm h})$ by empirical estimates $\widehat{R}^s({\bm h})$, $\widehat{R}^{t}_{u,C+1}({\bm h})$ and $\widehat{R}^s_{u,C+1}({\bm h})$ respectively, we have the following result.
\begin{corollary}\label{0.1}
Given a symmetric loss function $\ell$ satisfying the triangle inequality and bounded by $B$, and a hypothesis $\mathcal{H}\subset \{{\bm h}:\mathcal{X}\rightarrow \mathcal{Y}^t\} $  with conditions: 1) the constant vector value function ${\bm g}:={\mathbf{y}}_{C+1}\in \mathcal{H}$, 2) the Natarajan dimension of $\mathcal{H}$ is $d$, if a random labeled sample of size $n^s$ is generated by  source joint distribution ${P}_{X^sY^s}$-i.i.d. and a random unlabeled sample of size $n^t$ is generated by target marginal distribution $P_{X^t}$-i.i.d., then for any ${\bm h}\in \mathcal{H}$ and $\delta\in (0,1)$ with probability at least $1-3\delta$, we have
\begin{equation*}
\begin{split}
 \frac{R^t({\bm h})}{1-\pi_{C+1}^t}&\leq   {\widehat{R}^s({\bm h})}
 +{{2d_{\mathcal{H}}^{\ell}(P_{X^t|\mathcal{Y}^s},P_{X^s})}}+\widehat{\Delta}_o+\Lambda\\&+4B\sqrt{\frac{8d\log n^s+16d \log(C+1)+2\log{2/\delta}}{n^s}}\\&+2B\sqrt{\frac{8d\log n^t+16d \log(C+1)+2\log{2/\delta}}{{(1-\pi_{C+1}^t)}^2{n^t}}},
 \end{split}
\end{equation*}
where $\mathcal{X}$ is the feature space, $\mathcal{Y}^t$ is the target label space, ${\widehat{R}^s({\bm h})}$ is the empirical source risk, ${{R}^t({\bm h})}$ is the target risk, $\Lambda=\underset{{\bm h}\in \mathcal{H}}{\min}  ~R^s({\bm h})+R_*^t({\bm h})$ and empirical open set difference  $\widehat{\Delta}_o=\frac{\widehat{R}^t_{u,C+1}({\bm h})}{1-\pi_{C+1}^t}
 -\widehat{R}^s_{u,C+1}({\bm h})$, here $R^s({\bm h})$ are the risks defined in (\ref{kk}), $\widehat{R}^s_{u,C+1}({\bm h})$, $\widehat{R}^t_{u,C+1}({\bm h})$ are the empirical risks corresponding to ${R}^s_{u,C+1}({\bm h})$, ${R}^t_{u,C+1}({\bm h})$ defined in (\ref{10000}).
\end{corollary}
\begin{proof}
The proof is given in Appendix D.
\end{proof}

Next, we employ the open set difference $\Delta_o$ to construct our model --- Distribution Alignment with Open Difference.

  \subsection{{Algorithm}}
The importance of {Theorem \ref{-1000}} is that it tells us the relationships between the three terms (i.e., the source risk, the distribution discrepancy, and the open set difference) and the bound of the open set domain adaptation. Inspired by these relationships, our initial focus is  the following optimization problem for unsupervised open set domain adaptation:
\begin{equation}
  \begin{split}
    {\bm h}^*=\argmin_{{\bm h}\in \mathcal{H}} ~ &\widehat{R}^s({\bm h})+\lambda d^{\ell}_{\mathcal{H}}(\widehat{P}_{X^s},\widehat{P}_{X^t|\mathcal{Y}^s})
    \\
    &+\gamma \left(\frac{1}{1-\pi_{C+1}^t}\widehat{R}^t_{u,C+1}({\bm h})
 -\widehat{R}^s_{u,C+1}({\bm h})\right),
    \end{split}\label{allobje}
\end{equation}
where the hypothesis space $\mathcal{H}$ is defined as a subset of functional space $\{{\bm h}=[h_1,...,h_{C+1}]^T: h_c \in \mathcal{H}_k\}$ and $\lambda$ and $\gamma$ are two free hyper-parameters.

As proven by JDA \cite{long2013transfer}, ARTL \cite{DBLP:journals/tkde/LongWDPY14} and MEDA \cite{DBLP:conf/mm/WangFCYHY18}, incorporating conditional distributions into the original marginal distribution discrepancy can lead to superior domain adaptation performance. Hence, we have also added an additional conditional distribution discrepancy to the optimization problem in (\ref{allobje}). Hence, the new problem becomes:
\begin{equation*}
  \begin{split}
    {\bm h}^*=\argmin_{{\bm h}\in \mathcal{H}} ~ &\widehat{R}^s({\bm h})
    +\lambda \mu D_{{\bm h},k}^2(\widehat{P}_{X^s},\widehat{P}_{X^t|\mathcal{Y}^s})
    \\
    &+\lambda (1-\mu)\sum_{c=1}^C D_{{\bm h},k}^2(\widehat{P}_{X^s|{\mathbf{y}}_c},\widehat{P}_{X^t|{\mathbf{y}}_c})
    \\
    &+\gamma \left(\frac{1}{1-\pi_{C+1}^t}\widehat{R}^t_{u,C+1}({\bm h})
 -\widehat{R}^s_{u,C+1}({\bm h})\right),
    \end{split}
\end{equation*}
 where $\mu\in [0,1]$ is the adaptive factor \cite{DBLP:conf/mm/WangFCYHY18} to convexly combine the contributions from both the empirical marginal distribution alignment and the empirical conditional distribution alignment. Note that the original $d^{\ell}_{\mathcal{H}}(\cdot,\cdot)$ is replaced with the projected MMD $D_{{\bm h},k}(\cdot, \cdot)$ in the new problem, because $d^{\ell}_{\mathcal{H}}(\cdot,\cdot)$ is difficult to estimate. 
  This results in a gap with Theorem 1 where  the discrepancy distance is used to measure the distribution discrepancy rather than projected MMD. To mitigate this gap, we also give a similar theoretical bound using MMD distance (see Theorem 4 in Appendix C for details). Specifically, for proving Theorem 4, we need an additional condition that the loss $\ell$ is {\textit{squared loss}} $\ell(\mathbf{y},\mathbf{y}')=\|\mathbf{y}-\mathbf{y}'\|^2_2$. Thus, we use the squared loss to design our algorithm.

{Further, we have added the manifold regularization} \cite{DBLP:journals/jmlr/BelkinNS06} {to learn the geometric structure of the source and target domains. With this regularization, our algorithm can consistently achieve good performance when the setting degrades into a closed set domain adaptation (i.e., where there are no unknown classes). This is because the state of the art closed set algorithm ARTL} \cite{DBLP:journals/tkde/LongWDPY14} {is a special case of DAOD, when there is no open set difference.} 

Thus, the optimization problem can be rewritten as follows: 
\begin{equation}
  \begin{split}
    {\bm h}^*=\argmin_{{\bm h}\in \mathcal{H}} ~ &\widehat{R}^s({\bm h})
    +\lambda \mu D^2_{{\bm h},k}(\widehat{P}_{X^s},\widehat{P}_{X^t|\mathcal{Y}^s})
    \\
    &+\lambda (1-\mu)\sum_{c=1}^C D^2_{{\bm h},k}(\widehat{P}_{X^s|{\mathbf{y}}_c},\widehat{P}_{X^t|{\mathbf{y}}_c})
    \\
    &+\alpha\widehat{R}^t_{u,C+1}({\bm h})
 -\gamma \widehat{R}^s_{u,C+1}({\bm h})
 \\
 &+ \rho M_{\bm h}(\mathcal{S}_X,\mathcal{T}_X)+\sigma \|{\bm h}\|_k^2,
    \end{split}\label{4}
\end{equation}
where $\alpha := \gamma/(1-\pi^t_{C+1})$, $\rho$ and $\sigma$ are three free hyper-parameters, $\mathcal{T}_X$ denotes unlabeled target samples, $\mathcal{S}_X$ denotes source samples without labels, $M_{\bm h}(\mathcal{S}_X,\mathcal{T}_X)$ is the manifold regularization, and $\|{\bm h}\|_k^2$ is the squared norm of ${\bm h}$ in $\mathcal{H}_k$ to avoid over-fitting. 

Next, we show how to formulate equation (\ref{4}) using given samples. First, 
following the representer theorem, if the optimization problem $(\ref{4})$ has a minimizer ${\bm h}^*$, then ${\bm h}^*$ can be written as 
\begin{equation*}
{\bm h}^*(\mathbf{x})=\sum_{i=1}^{n^s+n^t}{\boldsymbol{\beta}}_ik(\mathbf{x}_i,\mathbf{x}),~~~\forall \mathbf{x} \in \mathcal{X}, 
\end{equation*}
where $\mathbf{x}_i\in \mathcal{S}_X\cup \mathcal{T}_X$ and ${\bm{\bm \beta}}_i \in \mathbb{R}^{{(C+1)}\times {1}}$ is the parameter. With this form of ${\bm h}^*$, we explain the computation of terms in $(\ref{4})$. The notations used in this section are summarized in Appendix A. Table II.

\subsubsection{Distribution Alignment} 

Since there are no labels for the target samples, we cannot directly compute 
\begin{equation}\label{2111}
\mu D_{{\bm h},k}^2(\widehat{P}_{X^s},\widehat{P}_{X^t|\mathcal{Y}^s})+(1-\mu)\sum_{c=1}^CD_{{\bm h},k}^2(\widehat{P}_{X^s|{\mathbf{y}}_c},\widehat{P}_{X^t|{\mathbf{y}}_c})
\end{equation}
Therefore, the pseudo target labels are used to help compute $\widehat{P}_{X^t|\mathcal{Y}^s}$ and $\widehat{P}_{X^t|{\mathbf{y}}_c}$ instead.
Given the pseudo target samples for known classes $\mathcal{T}_{X,K}$, the pseudo target samples for $c$-th class $\mathcal{T}_{X,c}$ and the source samples for $c$-th class $\mathcal{S}_{X,c}$
we can compute (\ref{2111}) by  the representer theorem and kernel trick as follows:
\begin{equation}\label{10}
{\rm tr}({\bm {\bm \beta}}^T\mathbf{KMK}{\bm {\bm \beta}}),
\end{equation}
where ${\bm {\bm \beta}}=[{\bm {\bm \beta}}_1,...,{\bm {\bm \beta}}_{n^s+n^t}]^T  \in \mathbb{R}^{{(C+1)}\times {(n^s+n^t)}}$, $\mathbf{K}$ is the ${(n^s+n^t)}\times{(n^s+n^t)}$ kernel matrix $[k(\mathbf{x}_i,\mathbf{x}_j)]$, here $\mathbf{x}_i,\mathbf{x}_j \in \mathcal{S}_X\cup \mathcal{T}_X$, and $\mathbf{M}=\mu \mathbf{M}_0+(1-\mu)\sum_{c=1}^{C}\mathbf{M}_c$ is the MMD matrix:

\begin{equation*}
(\mathbf{M}_0)_{ij}=\left\{
\begin{aligned}
&~~~~~\frac{1}{(n^s)^2},~~~\mathbf{x}_i, \mathbf{x}_j \in \mathcal{S}_X, \\
&~~~~~\frac{1}{(n^t_K)^2},~~~\mathbf{x}_i, \mathbf{x}_j\in \mathcal{T}_{X,K},\\
&~~~~~~~~~~~0,  ~~~\mathbf{x}_i~{\rm or}~\mathbf{x}_j\in \mathcal{T}_X \setminus \mathcal{T}_{X,K},\\
&~~~-\frac{1}{n^s{n}^t_K},~~\rm otherwise;
\end{aligned}
\right.
\end{equation*}

\begin{equation*}
(\mathbf{M}_c)_{ij}=\left\{
\begin{aligned}
&~~~\frac{1}{{(n^s_c)}^2},~~~ \mathbf{x}_i, \mathbf{x}_j \in\mathcal{S}_{X,c},\\
&~~~\frac{1}{{(n^t_c)}^2}, ~~~\mathbf{x}_i, \mathbf{x}_j\in \mathcal{T}_{X,c},\\
&-\frac{1}{n^s_c n^t_c},~~~\mathbf{x}_i \in\mathcal{S}_{X,c}, \mathbf{x}_j\in \mathcal{T}_{X,c},\\
&-\frac{1}{n^s_c n^t_c},~~~\mathbf{x}_j\in\mathcal{S}_{X,c}, \mathbf{x}_i\in \mathcal{T}_{X,c},\\
&~~~~~~~~~0,~~~\rm otherwise,
\end{aligned}
\right.
\end{equation*}
where $n^s:=|\mathcal{S}_X|$, $n^t_K:=|\mathcal{T}_{X,K}|$, $n^s_c:=|\mathcal{S}_{X,c}|$ and $n^t_c:=|\mathcal{T}_{X,c}|$.

\subsubsection{Manifold Regularization}
 The pair-wise affinity matrix is denoted as 
 \begin{equation*}
\mathbf{W}_{ij}=\left\{
\begin{aligned}
{\rm sim}(\mathbf{x}_i,\mathbf{x}_{j}), &~~~ \mathbf{x}_i \in \mathcal{N}_p(\mathbf{x}_{j})~or~\mathbf{x}_j\in \mathcal{N}_p(\mathbf{x}_i)  \\
0,  & ~~~\rm otherwise;
\end{aligned}
\right.
\end{equation*}
 where ${\rm sim}(\mathbf{x}_i,\mathbf{x}_{j})$ is the similarity function such as cosine similarity, $\mathcal{N}_p(\mathbf{x}_i)$ denotes the set of $p$-nearest neighbors to point $\mathbf{x}_i$ and $p$ is a free parameter.
The manifold regularization can then be evaluated as follows:
\begin{equation*}
\begin{split}
    M_{\bm h}(\mathcal{S}_X,\mathcal{T}_X)&=
    \sum^{n^s+n^t}_{i,j=1}\|{\bm h}(\mathbf{x}_i)-{\bm h}(\mathbf{x}_{j})\|_2^2\mathbf{W}_{ij}\\&=\sum_{c=1}^{C+1}\sum^{n^s+n^t}_{i,j=1}{ h}_c(\mathbf{x}_i)\mathbf{L}_{ij}{h}_c(\mathbf{x}_{j}),
    \end{split}
\end{equation*}
where $\mathbf{x}_{i},\mathbf{x}_{j}\in \mathcal{S}_X\cup \mathcal{T}_X$, $\mathbf{L}$ is the  Laplacian matrix, which can be written as $\mathbf{D-W}$, here $\mathbf{D}$ is a diagonal matrix and $\mathbf{D}_{ii}=\sum_{j=1}^{n^s+n^t}\mathbf{W}_{ij}$. Using the representer theorem and kernel trick, the manifold  regularization $M_{\bm h}(\mathcal{S}_X,\mathcal{T}_X)$ can be written as
\begin{equation}\label{11}
    {\rm tr}({\bm {\bm \beta}}^T\mathbf{KLK}{\bm {\bm \beta}}).
\end{equation}

\subsubsection{Open Set Loss Function} We use a matrix to rewrite the loss function and open set difference. Let the label matrix be 
$\mathbf{Y}\in \mathbb{R}^{(C+1)\times(n^s+n^t)}$:
\begin{equation}\label{Y}
\mathbf{Y}_{ij}=\left\{
\begin{aligned}
&~~~1,~~~ \mathbf{x}_j \in\mathcal{S}_{X,i} \\
&~~~0,~~~\rm otherwise
\end{aligned}
\right.
~~~{\rm when} ~~i\leq C,
\end{equation}
\begin{equation}\label{Y1}
~~~~~~\mathbf{Y}_{ij}=\left\{
\begin{aligned}
&~~~1,~~~ \mathbf{x}_j \in\mathcal{T}_{X,C+1} \\
&~~~0,~~~\rm otherwise
\end{aligned}
\right.
{\rm when} ~~i= C+1.
\end{equation}

 The  label matrix $\mathbf{\widetilde{Y}}\in \mathbb{R}^{(C+1)\times(n^s+n^t)}$ is
 \begin{equation}\label{Y2}
\mathbf{\widetilde{Y}}_{ij}= 1
~{\rm iff} ~i= C+1 {~\rm and}~\mathbf{x}_j \in\mathcal{S}_{X} , ~{\rm otherwise}~\mathbf{\widetilde{Y}}_{ij}=0.
\end{equation}

Then
\begin{equation}\label{12}
\begin{split}
    &{\widehat{R}^s({\bm h})} +\alpha\widehat{R}^t_{u,C+1}({\bm h})-\gamma\widehat{R}^s_{u,C+1}({\bm h})+\sigma \|{\bm h}\|_k^2\\=&{\|(\mathbf{Y}-{\bm {\bm \beta}}^T\mathbf{K})\mathbf{A}\|_F^2}-\gamma{\|(\mathbf{\widetilde{Y}}-{\bm {\bm \beta}}^T\mathbf{K})\mathbf{\widetilde{A}}\|_F^2}+\sigma {\rm tr}({\bm {\bm \beta}}^T\mathbf{ K}{\bm {\bm \beta}})
    \end{split}
\end{equation}
where $\mathbf{A}$ is a $(n^s+n^t)\times(n^s+n^t)$ diagonal matrix with $\mathbf{A}_{ii}=\sqrt{\frac{1}{n^s}}$ if $\mathbf{x}_i \in \mathcal{S}_X$,  $\mathbf{A}_{ii}=\sqrt{\frac{\alpha}{n^t}}$ if $\mathbf{x}_i\in \mathcal{T}_X$; $\mathbf{\widetilde{A}}$ is a $(n^s+n^t)\times(n^s+n^t)$ diagonal matrix with $\widetilde{\mathbf{A}}_{ii}=\sqrt{\frac{1 }{n^s}}$ if $\mathbf{x}_i \in \mathcal{S}_X$,  $\widetilde{\mathbf{A}}_{ii}=0$ if $\mathbf{x}_i\in \mathcal{T}_X$, and $\|\cdot\|_F$ is the Frobenius norm.

\subsubsection{Overall Reformulation}

Finally, based on (\ref{10}), (\ref{11}), (\ref{12}), the optimization problem in (\ref{4}) is reformulated as:
\begin{equation*}
 \begin{split}
{\bm \beta}^* = \argmin_{{\bm \beta} \in \mathbb{R}^{ (n^s+n^t)\times (C+1)}} \mathcal{L}({\bm \beta})
\end{split}
\end{equation*}
where 
\begin{equation}\label{HHH}
 \begin{split}
\mathcal{L}({\bm \beta}) :=  &{\|(\mathbf{Y}-{\bm {\bm \beta}}^T\mathbf{K})\mathbf{A}\|_F^2}-\gamma{\|(\mathbf{\widetilde{Y}}-{\bm {\bm \beta}}^T\mathbf{K})\mathbf{\widetilde{A}}\|_F^2} \\
&+ {\rm tr}(\mathbf{{\bm \beta}}^T\mathbf{K(\lambda M+\rho L) K}\mathbf{{\bm \beta}})
+\sigma {\rm tr}( \bm{\beta}^T\mathbf{ K}{\bm{\beta}}).
\end{split}
\end{equation}

\subsection{Training} 
There is a negative term in $\mathcal{L}({\bm \beta})$, hence it may be not correct to compute the optimizer by solving the equation $\frac{\partial \mathcal{L}({\bm \beta})}{\partial {\bm \beta}}=\mathbf{0}$ directly. Maybe the ``minimizer'' solved by $\frac{\partial \mathcal{L}({\bm \beta})}{\partial {\bm \beta}}=\mathbf{0}$ is a maximum point or a saddle point. Fortunately, the following {theorem} shows that there exists a unique optimizer that can be solved by $\frac{\partial \mathcal{L}({\bm \beta})}{\partial {\bm \beta}}=\mathbf{0}$.
\begin{theorem}\label{HH}
If the coefficient $\gamma$ of $\widehat{R}^s_{u,C+1}({\bm h})$  is smaller than $1$ and the kernel $k$ is universal, then the $\mathcal{L}({\bm \beta})$ defined in (\ref{HHH}) has a unique minimizer, which can be written as:
\begin{equation}\label{H1}
    \left((\mathbf{A}^2-\gamma\mathbf{\widetilde{A}}^2+\lambda \mathbf{M} +\rho \mathbf{L})\mathbf{K}+\sigma \mathbf{I}\right)^{-1}(\mathbf{A}^2\mathbf{Y}^T-\gamma\mathbf{\widetilde{A}}^2\mathbf{\widetilde{Y}}^T).
\end{equation}
\end{theorem}

The proof can be found in Appendix B.

 \begin{algorithm} [t]
     \caption{DAOD} 
      \KwIn{Data $\mathcal{S},\mathcal{T}_X$; $\#$iterations $T$; $\#$neighbor $p$ and parameters $\lambda,\sigma,\rho,\alpha, \gamma, \mu$; threshold $t$; universal kernel function $k(\cdot,\cdot)$.}
      
       1. $\widetilde{Y}_t\leftarrow ${ $\rm OSNN^{cv}$}$ (\mathcal{S},\mathcal{T}_X,t)$;\% Predict pseudo labels;\
       
       2. Compute $\mathbf{L,K}$ using $\mathcal{S}$, $\mathcal{T}_X$, and $\widetilde{Y}_t$;
       
       3. $i\leftarrow 1$\;

      \While{$i<T+1$} 
        { 
        4. Compute $\mathbf{M}$ using $\mathcal{S}$, $\mathcal{T}_X$, and $\widetilde{Y}_t$;
      
        5. Compute ${\bm \beta}$ by formula (\ref{H1})\;
        
        6. $\widetilde{Y}_{t}\leftarrow {\bm \beta}^T \mathbf{K}$;\%Predict pseudo labels\;
    
        7. $i\leftarrow i+1$\;
       } 
      \KwOut{Predicted target labels $\widetilde{Y}_t$, classifier ${\bm \beta}^T\mathbf{K}$.}
\end{algorithm}
 
To compute the true value of (\ref{H1}), it is best to use the groundtruth labels of the target domain.  However, our focus is on unsupervised task, which means that it is impossible to obtain any true target labels and, as mentioned, pseudo labels can be used instead. These pseudo labels are generated by applying an open set classifier that has been trained on the source samples to the target samples.
 
 In this paper, we used \textit{Open Set Nearest Neighbor for Class Verification-$t$} (${\rm OSNN^{cv}}$-$t$) \cite{DBLP:journals/ml/Mendes-JuniorSW17}  to help us learn the pseudo labels. We select the two nearest neighbors $\mathbf{v,u}$ from the test sample $\mathbf{s}$. 
 If both nearest neighbors have the
same label $\mathbf{y}_c$, $\mathbf{s}$ is classified with the label $\mathbf{y}_c$. Otherwise, the following ratio is calculated
 $  \|\mathbf{v}-\mathbf{s}\|_{2}/\|\mathbf{u}-\mathbf{s}\|_{2}$,
on the assumption that $\|\mathbf{v}-\mathbf{s}\|_{2}\leq\|\mathbf{u}-\mathbf{s}\|_2$. If the ratio is smaller than or equal to a pre-defined threshold $t$, $0<t<1$, $\mathbf{s}$ is classified with the same label as $\mathbf{v}$. Otherwise, $\mathbf{s}$ is recognized as the unknown sample.
 
To make the pseudo labels more accurate, we use the iterative pseudo label refinement strategy, proposed by JDA \cite{long2013transfer}. The implementation details are demonstrated in Algorithm 1 (https://github.com/fang-zhen/Open-set-domain-adaptation).

\section{Experiments and Evaluations}
In this section, we first utilized real world datasets to verify
the performance of DAOD. We then conducted experiments to examine the behavior of the parameters. 

\subsection{Real World Datasets}
We evaluated our algorithm on three cross-domain  recognition tasks: object recognition (\textbf{Office-31}, \textbf{Office-Home}), and face recognition (\textbf{PIE}). Table I lists the statistics of these datasets.
\begin{table}[h]
 \scriptsize
\caption{{Introduction of datasets.}}
\begin{center}
\begin{tabular}{p{1.3cm}lp{1cm}lp{1cm} lp{1cm}lp{1cm}lp{1cm}}
\hline \hline
Dataset  &Type&\#Sample& \#Feature&\#Class & Domain   \\ \hline
Office-31  &Object&~~4,110&~4,096&~~~31&~A,W,D \\ 
Office-Home  &Object&~15,500&~2,048&~~~65&Ar,Cl,Pr,Rw\\ 
PIE   &Face&~1,1554&~1,024&~~~68&~P1,...,P5\\
\hline
\hline
\end{tabular}
\end{center}
\end{table}

\textbf{Office-31} \cite{DBLP:conf/eccv/SaenkoKFD10} consists of three real-world object domains: AMAZON (\textbf{A}), DSLR (\textbf{D}) and WEBCAM (\textbf{W}). It has 4,652 images with 31 common categories. This means that there are 6 domain adaptation tasks: \textbf{A} $\rightarrow$ \textbf{D}, \textbf{A} $\rightarrow$ \textbf{W}, \textbf{D} $\rightarrow$ \textbf{A}, \textbf{W} $\rightarrow$ \textbf{A}, \textbf{D} $\rightarrow$ \textbf{W}, \textbf{W} $\rightarrow$ \textbf{D}. Following the standard protocol and for a fair comparison with the other algorithms, we extracted feature vectors from the fully connected layer-7 (fc7) of the AlexNet \cite{DBLP:conf/nips/KrizhevskySH12}.
We introduced an open set protocol for this dataset by taking classes 1-10 as the shared classes in alphabetical order. The classes 21-31 were used as the unknown classes in the target domain. 

\textbf{Office-Home} \cite{DBLP:conf/cvpr/VenkateswaraECP17} consists of 4 different domains: Artistic (\textbf{Ar}), Clipart (\textbf{Cl}), Product (\textbf{Pr}) and Real-World (\textbf{Rw}).  Each domain contains images from 65 object classes. We constructed 12 OSDA tasks: $\textbf{Ar}\rightarrow \textbf{Cl}$, $\textbf{Ar}\rightarrow \textbf{Pr}$,..., $\textbf{Rw}\rightarrow \textbf{Ar}$. In alphabetical order, we used the first 25 classes as the known classes and classes 26-65 as the unknown classes. Following the standard protocol and for a fair comparison with  the other  algorithms, we extracted feature vectors from ResNet-50. 

\textbf{PIE}\cite{DBLP:journals/pami/SimBB03} contains $41,368$ facial images of 68 people in various poses, illuminations, and expression changes.  The face images are captured by 13 synchronized cameras (different poses) and 21 flashes (different illuminations and/or expressions). We focused on 5 of 13 poses, i.e., \textbf{PIE1} (C05, left pose), \textbf{PIE2} (C07, upward pose), \textbf{PIE3} (C09, downward pose), \textbf{PIE4} (C27, frontal pose) and \textbf{PIE5} (C29, right pose).
These facial images were cropped to a size of $32\times32$. We took classes 1-20 as the known classes and classes 21-68 as the unknown classes in the target domain. 20 tasks were tested: \textbf{PIE1}$\rightarrow$\textbf{PIE2}, \textbf{PIE1}$\rightarrow$\textbf{PIE3},..., \textbf{PIE5}$\rightarrow$\textbf{PIE4}.

\begin{table*}[t] 
\centering
\caption{Acc(OS*) and Acc(OS) (\%) on \textbf{Office-31}, \textbf{Office-Home} and \textbf{PIE} Datasets. } 
\begin{tabular}{p{0.75cm}p{0.5cm}p{0.5cm}p{0.5cm}p{0.5cm}p{0.5cm}p{0.5cm}p{0.5cm}p{0.5cm}p{0.5cm}p{0.5cm}p{0.5cm}p{0.5cm}p{0.5cm}p{0.5cm}} 
\hline

Dataset&\multicolumn{2}{p{0.5cm}}{~~~~~${\rm {OSNN}}$}&\multicolumn{2}{p{0.5cm}}{~~~~~~TCA}&\multicolumn{2}{p{0.5cm}}{~~~~~~~JDA}&\multicolumn{2}{p{0.5cm}}{~~~~~~~JGSA}&\multicolumn{2}{p{0.5cm}}{~~~~~~~ATI}&\multicolumn{2}{p{0.5cm}}{~~~~~~OSBP}&\multicolumn{2}{p{0.5cm}}{~~~~~~DAOD}\\ 
\hline
&~OS*&~OS&~OS*&~OS&~OS*&~OS&~OS*&~OS&~OS*&~OS&~OS*&~OS&~OS*&~OS\\
\hline
A$\rightarrow$W  &56.0&54.0&54.8&54.8&63.0&64.8&75.7&75.2&70.6&69.7&69.1&70.1&\textbf{84.2}&\textbf{84.2}\\
A$\rightarrow$D &75.4&71.9&68.0&67.1&70.1&70.6&74.8&73.3&85.9&84.0&76.4&76.6&\textbf{89.8}&\textbf{88.5}\\
D$\rightarrow$A &62.6&60.3&53.4&52.7&60.4&60.7&62.4&61.5&68.3&67.6&62.3&62.5&\textbf{71.8}&\textbf{72.6}\\
D$\rightarrow$W &93.0&88.1&84.6&80.9&98.4&94.7&98.0&93.2&95.8&94.1&94.6&\textbf{98.9}&\textbf{98.0}&96.0\\
W$\rightarrow$A &58.6&56.8&56.1&55.6&62.5&62.6&64.0&62.9&64.0&62.8&\textbf{82.2}&\textbf{82.3}&72.9&74.2\\
W$\rightarrow$D &{99.3}&93.3&97.8&94.8&99.3&96.1&\textbf{100.0}&94.4&97.8&94.5&96.8&\textbf{96.9}&97.5&96.3\\
\hline
Average &74.2&70.7&69.2&68.5&75.6&74.9&79.2&76.7&80.4&78.8&80.2&80.4&\textbf{85.7}&\textbf{85.3}\\
\hline
\hline
Ar$\rightarrow$Pr &39.4&40.6&37.7&37.9&59.7&59.0&64.1&63.3&{70.4}&68.6&{69.2}&68.4&\textbf{72.6}&\textbf{71.8}\\
Ar$\rightarrow$Cl &32.1&33.7&24.4&24.1&39.1&39.6&45.9&46.0&54.2&53.1&53.3&53.1&\textbf{55.3}&\textbf{55.4}\\
Ar$\rightarrow$Rw &56.6&57.0&55.7&55.3&67.5&66.4&74.1&72.8&78.1&77.3&\textbf{79.1}&\textbf{78.0}&{78.2}&{77.6}\\
Cl$\rightarrow$Ar &32.3&34.0&31.3&32.1&41.9&42.1&43.8&44.5&59.1&57.8&58.2&57.9&\textbf{59.1}&\textbf{59.2}\\
Cl$\rightarrow$Pr &39.1&40.3&34.8&34.8&49.1&48.9&55.8&55.8&68.3&66.7&\textbf{72.4}&\textbf{71.6}&70.8&70.1\\
Cl$\rightarrow$Rw &46.9&47.7&41.4&41.2&59.7&59.1&62.8&62.5&75.3&74.3&72.3&71.4&\textbf{77.8}&\textbf{77.0}\\
Rw$\rightarrow$Ar &51.4&52.1&49.4&49.2&55.8&55.1&56.9&56.4&70.8&70.0&68.2&66.5&\textbf{71.3}&\textbf{70.5}\\
Rw$\rightarrow$Cl &38.0&39.2&34.9&34.1&44.1&43.9&48.7&48.6&55.4&55.2&\textbf{59.2}&\textbf{57.8}&{58.4}&\textbf{57.8}\\
Rw$\rightarrow$Pr &59.2&59.2&57.3&56.5&68.0&68.2&66.5&65.3&79.4&78.3&80.8&78.6&\textbf{81.8}&\textbf{80.6}\\
Pr$\rightarrow$Ar &38.5&39.7&33.2&33.4&48.4&48.0&55.8&55.5&62.6&61.2&61.0&59.6&\textbf{66.7}&\textbf{65.8}\\
Pr$\rightarrow$Cl &35.0&36.3&35.8&36.1&41.2&41.1&44.1&44.4&54.1&53.9&56.9&55.7&\textbf{60.0}&\textbf{59.1}\\
Pr$\rightarrow$Rw &59.6&59.7&58.3&57.5&70.4&68.9&73.5&72.3&81.1&79.9&83.9&82.1&\textbf{84.1}&\textbf{82.2}\\
\hline
Average &44.0&45.0&41.2&41.0&53.8&53.4&57.7&57.3&67.4&66.4&67.9&66.7&\textbf{69.6}&\textbf{68.9}\\
\hline
\hline
P1$\rightarrow$P2  &32.1&34.3&20.6&21.4&42.1&41.3&55.4&54.4&44.0&41.9&\textbf{66.6}&\textbf{64.2}&57.3&56.5\\
P1$\rightarrow$P3 &46.5&48.3&20.2&20.3&50.0&49.1&54.4&53.5&56.3&53.6&\textbf{69.1}&\textbf{66.4}&53.1&52.2\\
P1$\rightarrow$P4 &60.1&61.2&30.7&30.5&62.3&61.2&63.2&61.8&67.9&64.6&{80.0}&{76.2}&\textbf{85.2}&\textbf{82.4}\\
P1$\rightarrow$P5 &22.9&26.1&10.6&11.5&28.3&28.2&35.8&35.7&45.4&43.3&\textbf{50.2}&\textbf{49.1}&47.3&46.1\\
P2$\rightarrow$P1 &35.6&37.9&25.4&25.5&47.9&47.3&{68.5}&67.2&59.5&56.7&54.2&52.9&\textbf{69.7}&\textbf{68.1}\\
P2$\rightarrow$P3 &61.5&62.5&38.8&38.3&62.9&61.4&62.5&61.3&56.3&53.6&63.5&61.5&\textbf{71.7}&\textbf{69.9}\\
P2$\rightarrow$P4 &71.0&71.4&49.3&48.5&71.6&69.6&78.6&76.9&77.1&73.5&81.3&87.6&\textbf{91.2}&\textbf{88.2}\\
P2$\rightarrow$P5 &28.5&31.2&20.4&20.7&37.3&37.1&{49.0}&{48.0}&36.7&34.9&{44.2}&{41.2}&\textbf{49.8}&\textbf{49.4}\\
P3$\rightarrow$P1 &43.3&45.2&20.1&20.4&51.1&50.6&66.9&65.5&\textbf{68.4}&\textbf{66.9}&61.0&61.3&\textbf{68.3}&{66.6}\\
P3$\rightarrow$P2 &53.5&54.8&37.3&36.5&64.2&62.5&66.9&65.2&55.0&52.4&64.6&64.1&\textbf{70.4}&\textbf{68.5}\\
P3$\rightarrow$P4 &64.9&65.4&34.6&34.2&68.5&66.6&75.6&73.8&74.0&70.5&76.9&74.7&\textbf{87.1}&\textbf{83.9}\\
P3$\rightarrow$P5 &34.6&37.0&12.7&13.0&39.2&39.0&42.5&41.8&47.1&44.8&46.7&46.3&\textbf{53.3}&\textbf{52.3}\\
P4$\rightarrow$P1 &56.5&57.7&24.8&24.6&64.2&62.4&75.8&73.9&66.8&63.7&68.7&67.2&\textbf{87.1}&\textbf{84.4}\\
P4$\rightarrow$P2 &78.1&78.0&64.0&62.1&75.2&72.4&78.3&76.1&78.1&74.4&\textbf{85.0}&{82.2}&{84.8}&\textbf{82.4}\\
P4$\rightarrow$P3 &78.3&78.3&33.8&33.3&81.5&78.9&\textbf{81.3}&\textbf{79.1}&61.7&58.7&67.6&66.9&{80.0}&77.6\\
P4$\rightarrow$P5 &43.1&44.8&17.1&17.7&52.1&50.9&\textbf{65.8}&\textbf{64.4}&48.5&46.2&{63.8}&59.9&61.3&{59.9}\\
P5$\rightarrow$P1 &23.2&25.7&11.6&12.8&29.6&30.2&46.4&45.9&23.5&30.2&\textbf{66.6}&\textbf{64.2}&60.6&59.2\\
P5$\rightarrow$P2 &26.5&28.4&18.3&18.3&31.0&31.1&\textbf{44.0}&\textbf{43.6}&36.7&34.9&{35.8}&{35.4}&34.8&35.0\\
P5$\rightarrow$P3 &31.0&32.7&12.3&13.3&33.1&32.9&\textbf{55.4}&\textbf{54.6}&41.9&39.9&{46.3}&45.1&{44.4}&{44.6}\\
P5$\rightarrow$P4 &37.2&38.9&19.4&20.0&49.7&49.1&63.8&62.7&58.6&55.8&53.5&52.2&\textbf{70.3}&\textbf{68.6}\\
\hline
Average &46.4&48.0&26.2&26.1&52.1&51.1&61.5&60.3&55.2&53.0&62.2&61.0&\textbf{66.4}&\textbf{64.8}\\
\hline
\hline
All~avg&50.0&50.6&37.7&37.5&56.3&55.6&63.1&61.9&63.0&61.3&66.8&65.9&\textbf{70.4}&\textbf{69.3}\\
\hline
\end{tabular}
\end{table*}
\vspace{-0.2cm}
\subsection{Baseline Algorithms}
The baseline algorithms selected for comparison with DAOD were:

1) \textbf{No Transfer}:

$\bullet$ OSNN \cite{DBLP:journals/ml/Mendes-JuniorSW17}.  OSNN recognizes a sample as unknown by computing the ratio of similarity scores to the two most similar classes of the sample and then comparing the ratio with a pre-defined threshold.

2) \textbf{Closed Set}:

$\bullet$ TCA \cite{pan2011domain} + OSNN. The aim in implementing TCA is to show that if the UCSDA algorithm is used to solve the UOSDA problem, negative transfer will occur, leading to poor performance.

3) \textbf{Open Set}:

$\bullet$ JDA \cite{long2013transfer} + OSNN. We extended JDA into the open set setting. Joint distribution matching is the main step in JDA. Thus, we simply matched the known samples predicted by OSNN when the JDA algorithm was implemented.

$\bullet$ JGSA \cite{DBLP:conf/cvpr/ZhangLO17} + OSNN. We extended JGSA into the open set setting. First, for learning new features, we implemented JGSA using the source samples and the known target samples predicted by OSNN. Then, we used OSNN to predict the pseudo labels. We repeated the process until convergence. 

$\bullet$ ATI \cite{DBLP:conf/iccv/BustoG17} + OSNN. ATI was the first UOSDA algorithm, but it requires the unknown source samples to implement. Therefore, to implement ATI under our setting, we used ATI to select the outliers, and then learned the new features for matching the source domain and target domain excluding selected outliers. Lastly, OSNN was used to predict the labels.

$\bullet$ OSBP \cite{DBLP:conf/eccv/SaitoYUH18}. OSBP utilizes adversarial neural networks and a binary cross entropy loss to learn the probability for the target samples, then uses the estimated probability to recognize the unknown samples.

\subsection{{Experimental Setup}}
Before reporting the detailed evaluation results, it is important to
explain how DAOD's hyper-parameters are tuned. DAOD has several hyper-parameters: 1) the choice
of the kernel function $k$; 2) the adaptation parameters $\lambda,\sigma,\rho,p,\mu$; 3) the open set parameters $\alpha, \gamma$; and 4) $\#$iterations $T$ and the threshold $t\in (0,1)$.  Each parameter is discussed one by one next. 

\subsubsection{{The kernel function $k$} }

As suggested in \cite{gretton2012kernel,DBLP:conf/mm/WangFCYHY18}, we chose the Gaussian kernel
\begin{equation}
    k(\mathbf{a},\mathbf{b})=\exp(-\frac{\|\mathbf{a-b}\|^2_{2}}{2r^2}),
\end{equation}
where the kernel bandwidth $r$ is median$(\|\mathbf{a-b}\|_{2})$, $\forall \mathbf{a}, \mathbf{b}\in \mathcal{S}_X\cup\mathcal{T}_X$.

\subsubsection{{The adaptive factor $\mu$}}

The adaptive factor $\mu$ expresses the relative importance of the marginal distributions and conditional distributions. Wang et al. \cite{DBLP:conf/mm/WangFCYHY18} made the first attempt to compute $\mu$ by employing $\mathcal{A}$-distance \cite{DBLP:conf/nips/Ben-DavidBCP06}, which is the special case $d_{\mathcal{H}}^{0-1}$ of the discrepancy distance $d_{\mathcal{H}}^\ell$. According to \cite{DBLP:conf/nips/Ben-DavidBCP06}, the $\mathcal{A}$-distance can also be defined as the error of building
a binary classifier from hypothesis set $\mathcal{H}$ to discriminate between the two domains. Wang et al. \cite{DBLP:conf/mm/WangFCYHY18} used the linear hypothesis set to estimate $\mathcal{A}$-distance. Let $\epsilon({\bm h})$ be the error of the linear classifier $h$ discriminating source samples $\mathcal{S}_X$ and target samples $\mathcal{T}_X$. Then the $\mathcal{A}$-distance
\begin{equation*}
    d_{\mathcal{A}}(\mathcal{S}_X,\mathcal{T}_X)=2(1-\epsilon({\bm h})).
\end{equation*}
We adopted the same algorithm as \cite{DBLP:conf/mm/WangFCYHY18} to estimate $\mu$ by
\begin{equation*}
    \mu=1-\frac{d_0}{d_0+\sum_{c=1}^{C}d_c},
\end{equation*}
where $d_0:=d_{\mathcal{A}}(\mathcal{S}_X,\mathcal{T}_{X,K})$, $d_c:=d_{\mathcal{A}}(\mathcal{S}_{X,c},\mathcal{T}_{X,c}) ~(c=1,...,C)$. Here $\mathcal{T}_{X,K}$ is the set of the target samples predicted as known samples. This estimation has to be computed at every iteration of DAOD, since the predicted conditional distributions for the target may
vary each time.
\subsubsection{{The open set parameters $\alpha$ and $\gamma$}}

As shown in Figs. 3 and 4, DAOD is able to achieve consistently good performance within a same range $\alpha \in [0.2, 0.4]$ and $\gamma \in [0.15, 0.5]$, which shows the relative stability of DAOD given the correct tuning of these two parameters.
Tuning should be done according to the following rules. First, the positive term  $R^t_{u,C+1}$ and the negative term $R^s_{u,C+1}$ in the open set difference are inferred from each other. A larger positive term means that more samples are recognized as the unknown classes. A larger negative term implies that more samples are classified as known classes. To ensure that the positive and negative terms balance, the difference $|\alpha-\gamma|$ should not be too large.  Further, the parameter $\alpha$ should be larger than $\gamma$, since the positive term's coefficient  $1/(1-\pi^t_{C+1})$ is larger than $1$. In this paper, we set $\alpha=0.4$ for all tasks and 1) $\gamma=0.2$ for \textbf{Office-31}, and 2) $\gamma=0.25$ for \textbf{Office-Home} and \textbf{PIE} datasets.

\subsubsection{{Other hyper-parameters}}
We ran DAOD with a wide range of parameter values for $\lambda$, $\rho$, $p$, $\sigma$, $t$ and $T$ in Section V-G. The results are shown in Fig. 4. These results indicate that DAOD can provide a robust performance with a wide range of hyper-parameter values.

From our tests, the best choices of parameters were: $\lambda \in [50,100]$, $\rho \in [0,1]$, $p\in [2,32]$, $\sigma \in [0.2,1.6]$, $t\in [0,0.9]$ and DAOD can converge within $10$ iterations. To sum up, the performance of DAOD stays robust with a large range of parameter choice. Therefore, the parameters do not need to be significantly fine-tuned in practical applications.
In this paper, we fixed $p=10$, $\rho=1$ and $\sigma=1$, $T=10$, $t=0.5$ and set 1) $\lambda=50$ for \textbf{Office-31}, and 2) $\lambda=500$ for \textbf{Office-Home} and \textbf{PIE} datasets.

Although DAOD is easy to use, and its parameters do not have to be fine-tuned, we did explore how to further tune these parameters for research purposes. We chose the parameters according to following the rules: 1) The regularization term $\|{\bm h}\|^2_k$ is very important, so we tended to choose a slightly larger $\sigma$ ($\sigma=1$) to prevent DAOD from degenerating. 2)  We chose $\rho$ by following \cite{DBLP:journals/jmlr/BelkinNS06}. 3) $p$ is set following \cite{DBLP:journals/jmlr/BelkinNS06,DBLP:journals/pami/CaiHHH11}. 4)  distribution alignment is inevitable for DAOD, so we chose a larger $\lambda$ $(\lambda\geq 50)$ to make it count.

\begin{figure*}[t]
\centering
\includegraphics[scale=0.4,trim=90 30 50 0, clip]{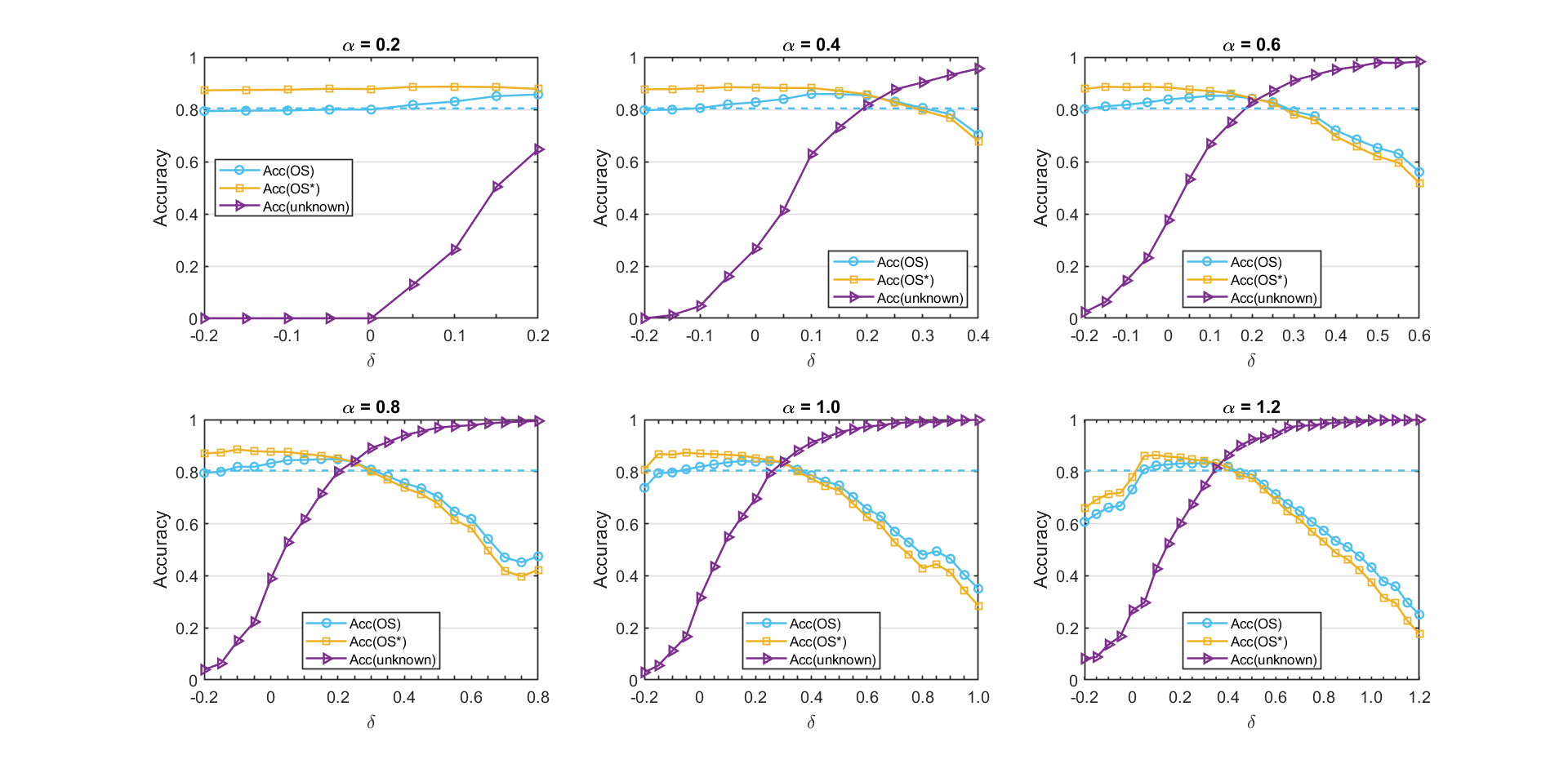}
\caption{The horizontal axis is the difference in the open set parameters $\delta=\alpha-\gamma$. In the figures, the difference $\delta$ is not  larger than  $\alpha$, since the parameter $\gamma$ is required to be larger than or equal to $0$. If $\delta>0$, $\alpha$ is larger than $\gamma$. If $\delta<0$, $\gamma$ is larger.}
\end{figure*}

We used two  types of accuracy \cite{DBLP:conf/iccv/BustoG17,DBLP:conf/eccv/SaitoYUH18} to evaluate DAOD:
\vspace{-0.cm}
\begin{equation}\label{100}
    {\rm Acc{(OS)}}=\frac{1}{C+1}\sum_{c=1}^{C+1}\frac{|x:x~{\rm from~class~} c \bigwedge \widetilde{f}(\mathbf{x})=c|}{|x:x ~{\rm from ~class~ } c |},
\end{equation}
and
\begin{equation}\label{101}
    {\rm Acc{(OS^*)}}=\frac{1}{C}\sum_{c=1}^{C}\frac{|x:x~{\rm from~class~} c \bigwedge \widetilde{f}(\mathbf{x})=c|}{|x:x ~{\rm from ~class~ } c |},
\end{equation}
where  $\widetilde{f}$ is the predicted classifier. Note that Acc(OS) is the main index for evaluating the performance of the UOSDA algorithms\cite{DBLP:conf/iccv/BustoG17}.
\subsection{Experimental Results}
The classification accuracy of the UOSDA tasks is shown in Table II. The following facts can be observed from this table. 1) The closed set algorithm TCA performed poorly on most tasks, even worse than the standard OSNN algorithm, indicating that negative transfer occurred. 2) All open set algorithms achieved better classification accuracy than
OSNN on most tasks. This is because the source samples and the known target samples have different distributions. 3) DAOD achieved much better performance ${\rm Acc(OS)}$ than the six baseline algorithms on most tasks ($24$ out of $38$). The average classification accuracy (${\rm Acc(OS),Acc(OS^*)}$) of DAOD on the $38$ tasks was $69.3\%$, $70.4\%$ respectively, gaining a  performance improvement of $3.4\%$, $3.6\%$ compared to the best baseline OSBP. 4) Generally, JDA+OSNN, JGSA+OSNN and ATI+OSNN algorithms did not perform as well as DAOD. A major limitation of these algorithms may be that they omit the selected unknown target samples when they construct a latent space to match the distributions for the known classes. This may result in the  unknown samples being mixed with the known samples in the latent space. In DAOD, the negative term $R^s_{u,C+1}$ helps DAOD to avoid the problem suffered by JDA, JGSA and ATI. 5) The performance of the OSPB algorithm was generally worse than that of DAOD. The main reasons may be that: 1) OSBP only matches the marginal distributions, not the joint distributions; 2) OSBP does not keep the unknown target samples away from the known source samples, with the result that many unknown target samples are recognized as known samples. DAOD, however, uses the negative term $R^s_{u,C+1}$ to separate the source samples and unknown target samples.

\begin{figure*}[h]
\centering
\includegraphics[scale=0.45,trim=145.7 30 30 37, clip]{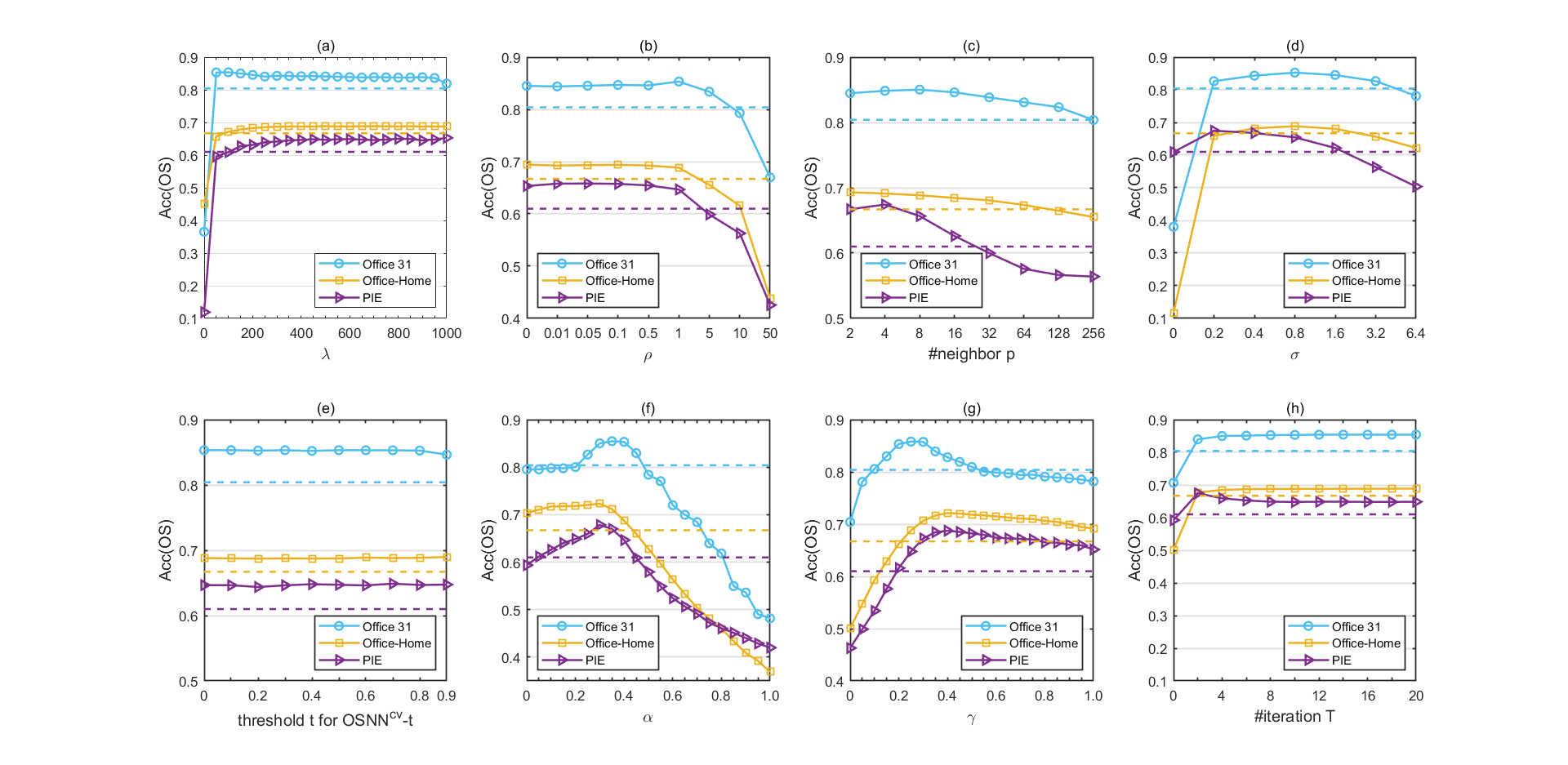}
~~~~~~~~\caption{Parameter sensitivity study, ablation study and convergence analysis of the proposed DAOD algorithm.}
\end{figure*}

\subsection{Open Set Parameters Analysis} From our analysis of the open set parameters $\alpha$ and $\gamma$, we find that the relationship between $\alpha$ and $\gamma$ is closely related to another parameter, the difference $\delta:=\alpha-\gamma$. We conducted experiments on the \textbf{Office-31} dataset with $\alpha$ ranging from $0.2$ to $1.2$ and $\delta$ ranging from $-0.2$ to $\alpha$. Due to space limitations, the average results on \textbf{Office-31} are reported in Fig. 3. According to Fig. 3, we made the following observations:

1) As $\delta$ increased, the accuracy of the unknown classes also increased, since a larger positive term $R^t_{u,C+1}({\bm h})$ means that more samples are recognized as unknown.

2) When $\delta<0$ ($\alpha<\gamma$), for almost all $\alpha\in [0.2,1.2]$, the performance Acc(OS) was poorer than the best baseline algorithm (dashed line). This is because when $\delta<0$, more samples are recognized as known classes.  Our theoretical results support this observation (Theorem 1) since the positive term’s coefficient $1/(1-\pi_{C+1}^t)$ is larger than the negative term's coefficient $1$. Thus, $\delta$ should be larger than $0$ ($\alpha>\gamma$).

3) All figures in Fig. 3 are similar for almost all $\alpha$ from $0.4$ to $1.2$,  which implies that  $\alpha$ may be not the most important factor influencing the performance of DAOD. Rather, the difference $\delta$ is likely to be more important.

4)	Performance Acc(OS) begins to decrease when $\delta$ is larger than $0.25$ because more known samples are classified as unknown with a larger $\delta$.

5) When $\alpha$ ranged between $0.2$ to $1.2$ and $\delta$ was chosen from $[0.05,0.2]$, the performance Acc(OS) of DAOD $\delta$ was superior to the best baseline.

6) Although $\alpha$ is not the main factor influencing the performance of DAOD, we compare figures ($\alpha<1.0$) with figures ($\alpha\geq1.0$) and find that a smaller $\alpha$ achieves slightly better performance than a larger $\alpha$. In general, we select $\alpha$ from  $[0.2,0.4]$ and $\delta$ from $[0.05,0.25]$.

 \subsection{{Parameter Sensitivity, Ablation Study and Convergence Analysis. }}
 
 These studies were conducted on different types of datasets to demonstrate that: 1) a wide range of parameter values can be chosen to obtain satisfactory performance, and 2) the open set difference and distribution alignment term are important and necessary.   We evaluated important parameters $\lambda, \sigma, \rho$, $p$, $t$, $\alpha$, $\gamma$ and $T$, reporting the average results for datasets \textbf{Office-31}, \textbf{Office-Home} and \textbf{PIE} respectively. The dashed line denotes the results of the best baseline algorithm with each dataset.

 \textbf{{Distribution Alignment $\lambda$.}} We ran DAOD with varying
values of $\lambda$.  Fig. 4(a) plots the
classification accuracy w.r.t. to different values of $\lambda$. From this figure, we can see that: 1) When $\lambda=0$, the performance was the worst than the baseline. 2) After the increasing of the $\lambda$ from $0$ to $50$, the performance dramatically increased to equal that of the baseline. 3) From $50$ to $1000$, DAOD was stable with values of around $0.85, 0.7$, and $0.65$ on the three datasets. Overall, the performance of DAOD with most values of $\lambda$ was better than the baselines.  We also found that larger values of $\lambda$ resulted in a better distribution alignment, and, if we chose $\lambda$ from $[50,1000]$, we obtained better results than the best baseline algorithm.

\textbf{{Manifold Regularization $\rho$.}} In these experiments, we ran DAOD with varying values of $\rho$. Larger values of $\rho$ increase the importance of manifold consistency in DAOD.  From Fig. 4(b), we can see that: 1)  DAOD's performance was steady and consistently good when $\rho\in [0,1]$. 2) But, after the increasing of $\rho$ from $1$ to $5$,  its performance dramatically dropped below the baseline. 3) Further, DAOD's continued to fall below the baseline from $5$ to $50$.   The reason for this poor performance at $\rho\in [5,50]$ is that when $\rho$ is large, DAOD mainly focuses on the geometric information of the samples and ignores other information.  Choosing $\lambda$ from $[0,1]$, however, provides the best results.

\textbf{{\#Nearest Neighbors $p$.}} We ran DAOD with varying values of $p$. If $p\to +\infty$, two samples which are not at all similar are connected. If $p\to 0$, limited similarity information between samples is captured, thus $p$ should not be too large or too small. Fig. 4(c) shows that if $p$ is selected from $[2,32]$, the performance of our algorithm is better than the baseline. When $p>32$, the performance of $\textbf{PIE}$ was worse than the baseline. One reason may be that when $p$ is large, the samples from different classes are connected, resulting in that samples from different classes share similar scores. From Fig. 4(c), $p$ can be selected from $[2,32]$.

\textbf{{Regularization $\sigma$.}} We ran DAOD with varying values of $\sigma$ and plotted the
classification accuracy as shown in Fig. 4(d). Theoretically, when $\sigma\to 0$, the classifier
degenerates and overfitting occurs. When $\sigma\to +\infty$, the classifier obtains a trivial result. From Fig. 4(d),  we  can  see  that: 1) When $\sigma=0$,  the  performance  was  the  worst  and  also  much worse than the baseline. 2) However, after increasing of $\sigma$ from $0$ to $0.2$, performance dramatically increased commensurate with the baseline. 3) From $0.2$ to $1.6$, DAOD was stable with values at around $0.85, 0.7$  and $0.65$ on  three  datasets. 4) When $\sigma>1.6$, the performance dramatically dropped again to below the baseline. According to Fig. 4(d), we can choose $\sigma\in[0.2,1.6]$.

\textbf{{Threshold $t$.}} Fig. 4(b) shows the classification accuracy with varying values of $t$. Theoretically, the threshold $t$ is determined by the openness $\mathbb{O}$. When openness $\mathbb{O}\to 1$, $t\to 0$. When  openness $\mathbb{O}\to 0$, $t\to 1$. However, according to Fig. 4(e), DAOD performed steadily when the threshold $t$ varies from $[0,0.9]$. This is because: 1) As the number of iterations $T$ increases, the effect of $t$ tapers off. 2) ${\rm OSNN^{cv}}$-$t$ is not sensitive to $t$.

{\textbf{Open Set Parameter $\alpha$.}} Fig. 4(f) plots the classification accuracy w.r.t. different values. Theoretically, when $\alpha \to 0$, the classifier can not recognize unknown samples, whereas, when $\alpha \to +\infty$, the classifier classifies all samples as unknown. These conjectures are verified by the results in Fig. 4(f), where the performance reaches its maximal point at around $\alpha=0.3$ and then gradually drops as $\alpha$ increases. Performance was worst and lower than the baselines when $\alpha>0.4$, because at this parameter setting many samples from known classes are classified as unknown. In general, we can choose $\alpha$ from $[0.2,0.4]$.

{\textbf{Open Set Parameter $\gamma$.}} The classification accuracy w.r.t. different values of $\gamma$ is shown in Fig. 4(g). Theoretically, when $\gamma \to +\infty$, DAOD keeps the unknown target samples away from known source samples. As a result, few samples are classified as unknown classes. When $\gamma \to 0$, more samples are classified as unknown, and when $\gamma < 0.15$, its performance was worse than the baselines. Conversely, as $\gamma$ increased, DAOD's performance dramatically increased, reaching its maximal value at around $\gamma=0.3$ before gradually dropping again as $\gamma$ continues to increase. In general, we can choose $\gamma\in[0.15,0.5]$.

{\textbf{Ablation Study.} } 
1) $\alpha$ and $\gamma$ are the two parameters that control the contribution of the open set difference. As shown in Fig. 4(f), setting $\alpha$ closer to $0$ reduces the contribution of the open set difference and performance degrades compared to the optimal value of about $\alpha = 0.3$. Further, as shown in Fig. 4(g), setting $\gamma$ closer to $0$ also reduces the contribution of the open set difference and again performance degrades compared to the optimal value of about $\gamma=0.3$. Therefore, we can safely draw the conclusion that our proposed open set difference is a necessary term for open set domain adaption. 2) $\lambda$ is the parameter that controls the contribution of the distribution discrepancy. As shown in Fig. 4(a), when $\lambda$ is $0$, performance is much worse than at other values, which shows that this term also makes a significant contribution to the final domain adaptation performance. 3) {$\rho$ controls the contribution of the manifold regularization. Fig. 4(b) shows there is no significant change in performance when $\rho$ is set in the range $0$ to $1$. These results indicate that $\rho$ makes no significant  contributions to DAOD and may even negatively effect its performance with values from $5$ to $50$. { Though the contribution of  manifold regularization is not significant,  more experiments in Appendix E show that the manifold regularization is necessary}}. 4) $\sigma$ is used to avoid overfitting. As shown in Fig. 4(d), performance drops significantly when $\sigma$ is set to 0. Thus, the term $\|{\bm h}\|_k^2$ is important to our algorithm.

\textbf{Convergence Analysis.}  The results of the convergence analysis on the number of iterations $T$ are provided in Fig. 4(h). As shown, DAOD reached a steady performance in only a few iterations $(T < 10)$. This is a clear indication of the advantages of DAOD's ability to be trained in unsupervised open set domain adaptation tasks.

\section{Conclusion and Future Work }
To the best of our knowledge, this is the first work to present a theoretical analysis for open set domain adaptation. In deriving a theoretical bound, we discovered a special term, open set difference, which is crucial for recognizing unknown target samples. Using this open set difference, we then constructed an unsupervised open set domain adaptation algorithm, called Distribution Alignment with Open Difference (DAOD). Extensive experiments show that DAOD outperforms several competitive algorithms.

In the future, we will mainly focus on \textit{universal domain adaptation} \cite{longmingsheng1}, which is a unified domain adaptation framework that includes closed set domain adaptation, open set domain adaptation and partial domain adaptation \cite{DBLP:conf/cvpr/CaoL0J18} .


%


\section*{Acknowledgment}

The  work  presented  in  this  paper  was  supported  by  the Australian Research Council (ARC) under DP170101632 and FL190100149.  We also wish to thank the anonymous reviewers for their helpful comments.

\ifCLASSOPTIONcaptionsoff
  \newpage
\fi



%

\bibliographystyle{IEEEtran}
\bibliography{mybibfile}
\vspace{-1.4cm}
\begin{IEEEbiography}[{\includegraphics[width=1.0in,height=1.8in,clip,keepaspectratio]{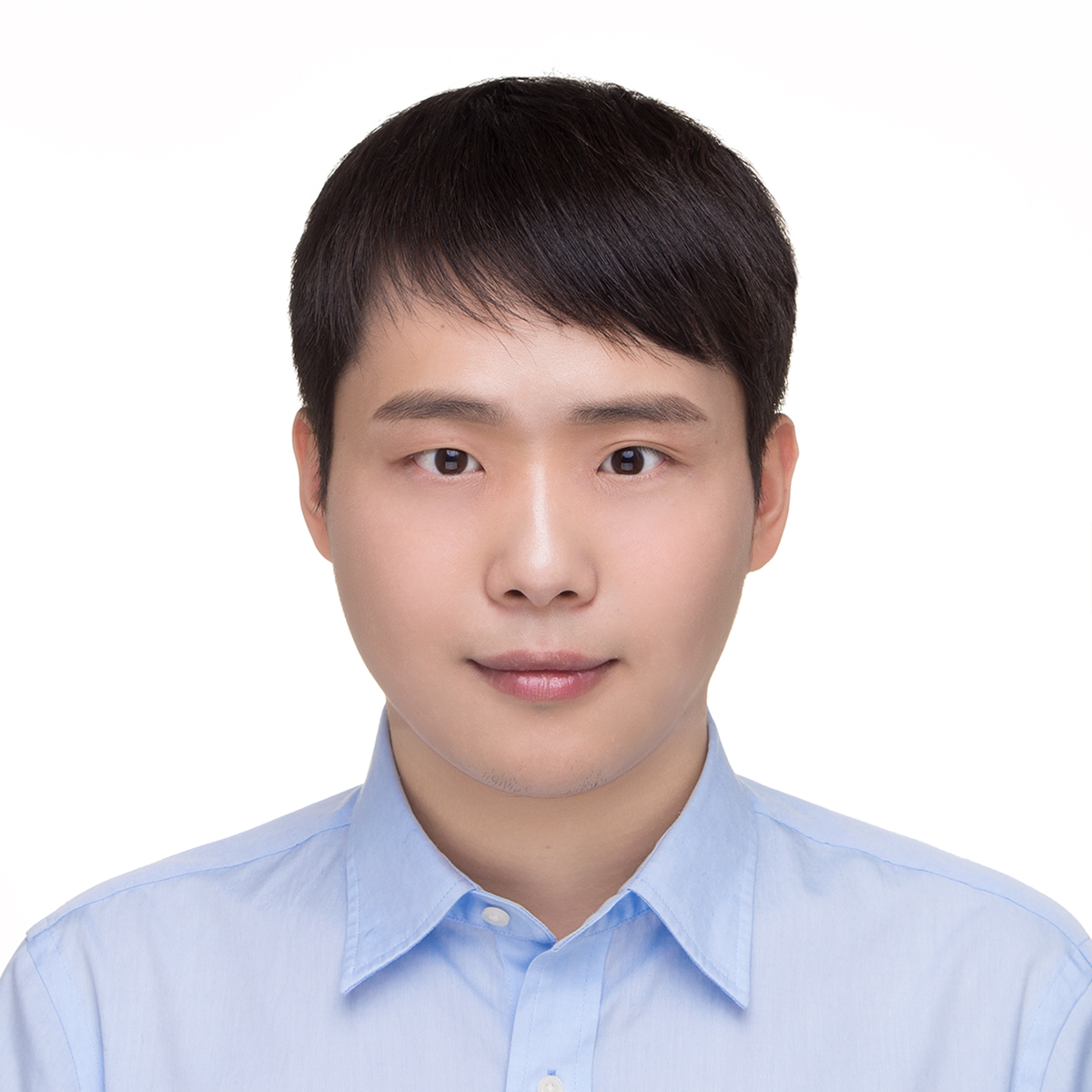}}]{Zhen Fang}
received his M.Sc. degree in pure mathematics from the School of Mathematical Sciences Xiamen University, Xiamen, China, in 2017. He is working toward a PhD degree with the Faculty of Engineering and Information Technology, University of Technology Sydney, Australia. His research interests include transfer learning and domain adaptation. He is a Member of the Decision Systems and e-Service Intelligence (DeSI) Research Laboratory, CAI, University of Technology Sydney.
\end{IEEEbiography}

\begin{IEEEbiography}[{\includegraphics[width=1in,height=1.25in,clip,keepaspectratio]{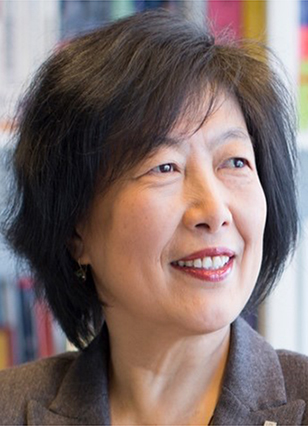}}]{Jie Lu} (F’18) is a Distinguished Professor and the Director of the Centre for Artificial Intelligence at the University of Technology Sydney, Australia. She received her PhD degree from Curtin University of Technology, Australia, in 2000. Her main research interests arein the areas of fuzzy transfer learning, concept drift, decision support systems, and recommender systems. She is an IEEE fellow, IFSA fellow and Australian Laureate fellow. She has published six research books and over 450 papers in refereed journals and conference proceedings; has won over 20 ARC Laureate, ARC Discovery Projects, government and industry projects. She serves as Editor-In-Chief for Knowledge-Based Systems (Elsevier) and Editor-In-Chief for International journal of computational intelligence systems. She has delivered over 25 keynote speeches at international conferences and chaired 15 international conferences. She has received various awards such as the UTS Medal for Research and Teaching Integration (2010), the UTS Medal for Research Excellence (2019), the Computer Journal Wilkes Award (2018), the IEEE Transactions on Fuzzy Systems Outstanding Paper Award (2019), and the Australian Most Innovative Engineer Award (2019).
\end{IEEEbiography}

\begin{IEEEbiography}[{\includegraphics[width=1in,height=1.25in,clip,keepaspectratio]{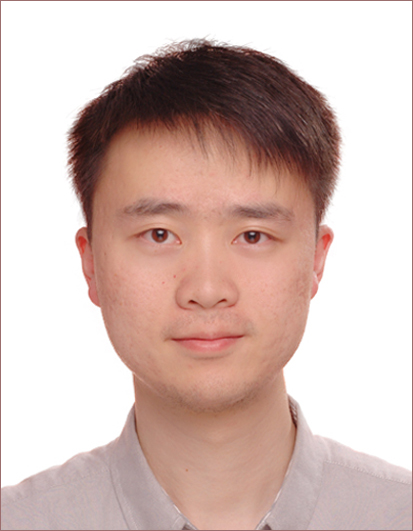}}]{Feng Liu}
is a Doctoral candidate in Centre for Artificial intelligence, Faculty of Engineering and Information Technology, University of Technology Sydney, Australia. He received an M.Sc. degree in probability and statistics and a B.Sc. degree in pure mathematics from the School of Mathematics and Statistics, Lanzhou University, China, in 2015 and 2013, respectively. His research interests include domain adaptation and two-sample test. He has served as a senior program committee member for ECAI and program committee members for NeurIPS, ICML, IJCAI, CIKM, FUZZ-IEEE, IJCNN and ISKE. He also serves as reviewers for TPAMI, TNNLS, TFS and TCYB. He has received the UTS-FEIT HDR Research Excellence Award (2019), Best Student Paper Award of FUZZ-IEEE (2019) and UTS Research Publication Award (2018).
\end{IEEEbiography}

\begin{IEEEbiography}[{\includegraphics[width=1in,height=1.25in,clip,keepaspectratio]{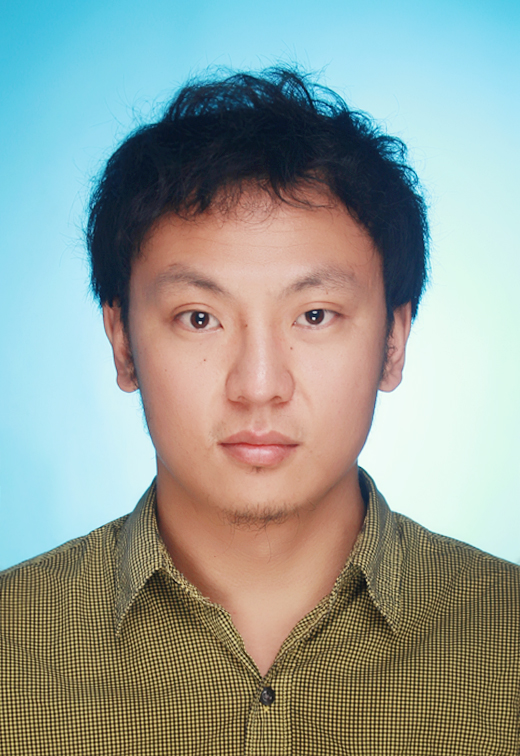}}]{Junyu Xuan}
is working as a Postdoctoral Research Fellow in the Faculty of Engineering and IT at the University of Technology Sydney in Australia. His main research interests include Machine Learning, Bayesian Nonparametric Learning, Text Mining, and Web Mining. He has published almost 40 papers in high-quality journals and conferences, including Artificial Intelligence, Machine Learning, IEEE TNNLS, ACM Computing Surveys, IEEE TKDE, ACM TOIS, and IEEE TCYB, etc.
\end{IEEEbiography}

\vspace{-0.5cm}
\begin{IEEEbiography}[{\includegraphics[width=1in,height=1.25in,clip,keepaspectratio]{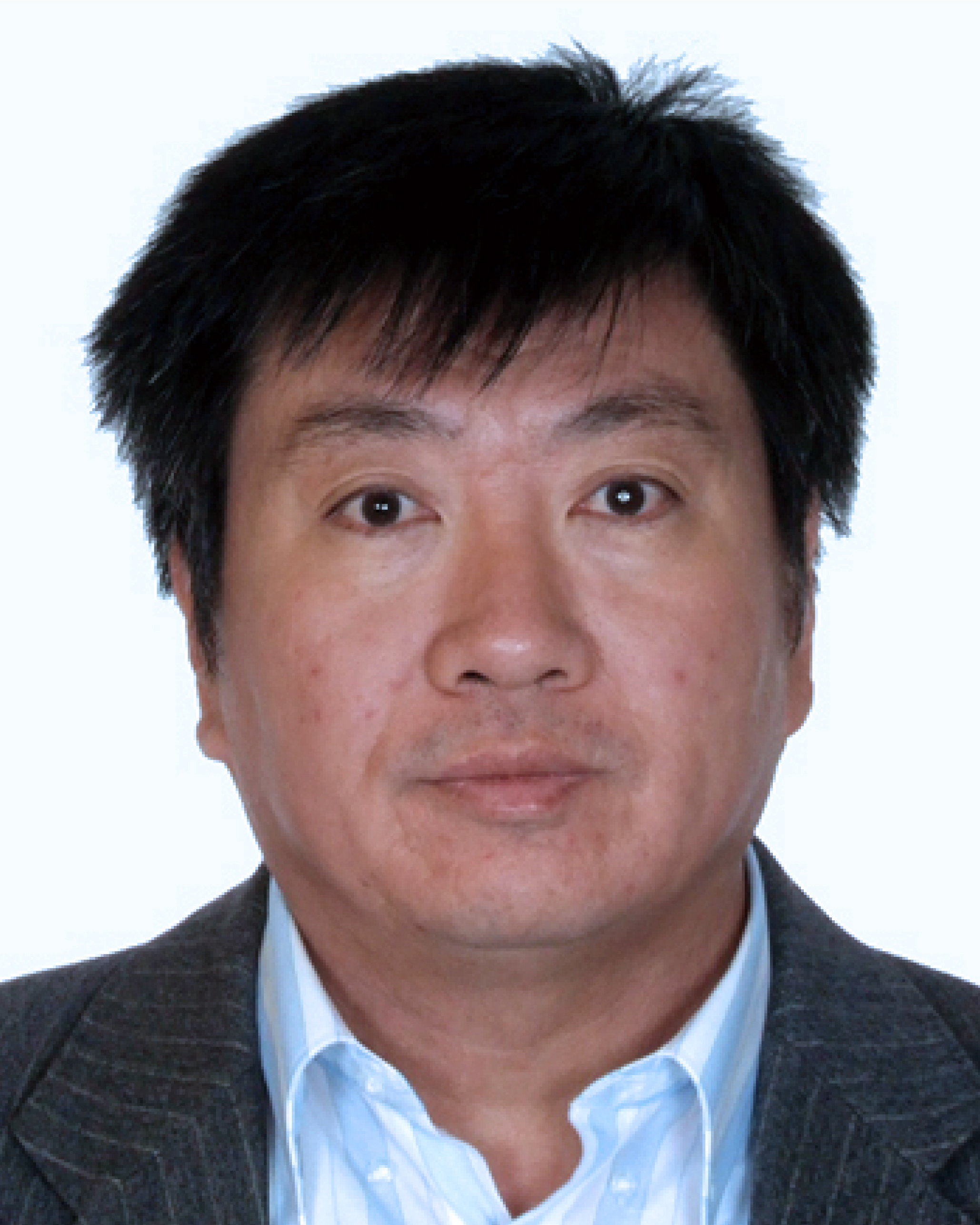}}]{Guangquan Zhang}
is a Professor and Director of the Decision Systems and e-Service Intelligent (DeSI) Research Laboratory, Faculty of Engineering and Information Technology, University of Technology Sydney, Australia. He received his PhD in applied mathematics from Curtin University of Technology, Australia, in 2001.
His research interests include fuzzy machine learning, fuzzy optimization, and machine learning and data analytics. He has authored four monographs, five textbooks, and 350 papers including 160 refereed international journal papers. Dr. Zhang has won seven Australian Research Council (ARC) Discovery Project grants and many other research grants. He was awarded an ARC QEII Fellowship in 2005. He has served as a member of the editorial boards of several international journals, as a guest editor of eight special issues for IEEE Transactions and other international journals, and has co-chaired several international conferences and work-shops in the area of fuzzy decision-making and knowledge engineering. 
\end{IEEEbiography}

\newpage
\newpage

\includepdf[pages=1-22]{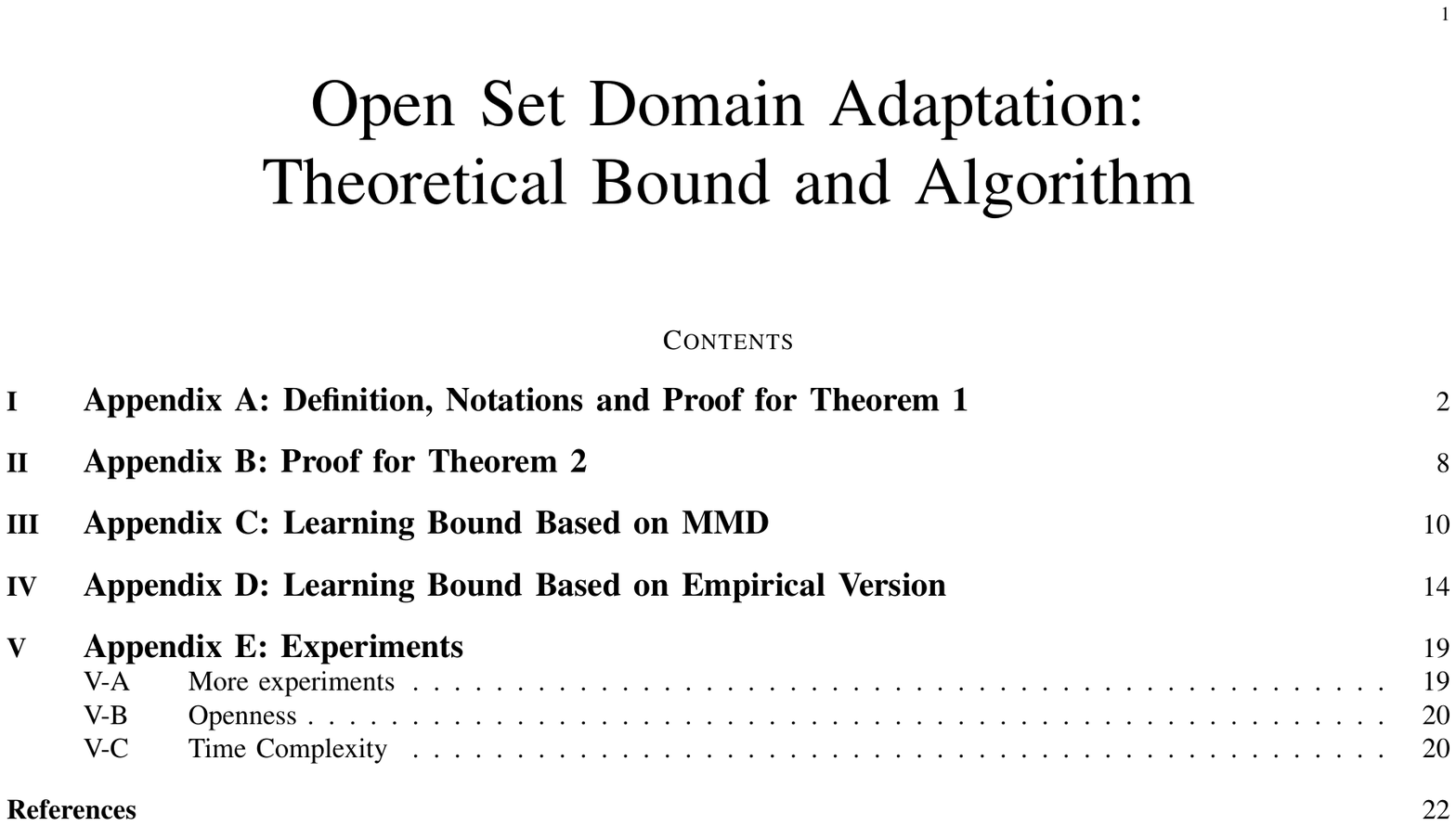}

%








\end{document}